%% file: main.tex
\documentclass{elsarticle}

\usepackage{times}
\usepackage{latexsym}
\usepackage{graphicx}
\usepackage{subcaption}

\usepackage[T1]{fontenc}

\usepackage[utf8]{inputenc}

\usepackage{microtype}

\usepackage{array}
\usepackage{pifont}
\usepackage{tabularx}
\usepackage{adjustbox}
\usepackage{multirow}
\usepackage{enumitem}
\usepackage{xspace}
\usepackage{tcolorbox}
\usepackage{booktabs,amsfonts,dcolumn}
\usepackage{hyperref}
\usepackage{url}
\usepackage{amsmath,amsthm,amsfonts,amssymb,bm,stmaryrd,bbm}
\usepackage[noorphans,vskip=0.75ex,leftmargin=2ex]{quoting}

\usepackage{xcolor}

\hypersetup{
pdftitle={Healthcare Language Model Embedding Spaces},
pdfauthor={Niall Taylor},
}

\title{Developing Healthcare Language Model Embedding Spaces}

\begin{document}

\begin{frontmatter}

\author[inst1]{Niall Taylor}
\author[inst2]{Dan Schofield}
\author[inst1]{Andrey Kormilitzin}
\author[inst3]{Dan W Joyce}
\author[inst1]{Alejo Nevado-Holgado}
\affiliation[inst1]{organization={Department of Psychiatry, University of Oxford},
            city={Oxford},
            country={United Kingdom}}

\affiliation[inst2]{organization={Transformation Directorate, NHS England},
            city={Leeds},
            country={United Kingdom}}

\affiliation[inst3]{organization={Department of Primary Care and Mental Health, University of Liverpool},             
            city={Liverpool},
            country={United Kingdom}}

\begin{abstract}
Pre-trained Large Language Models (LLMs) often struggle on out-of-domain datasets like healthcare focused text. We explore specialized pre-training to adapt smaller LLMs to different healthcare datasets. Three methods are assessed: traditional masked language modeling, Deep Contrastive Learning for Unsupervised Textual Representations (DeCLUTR), and a novel pre-training objective utilizing metadata categories from the healthcare settings. These schemes are evaluated on downstream document classification tasks for each dataset, with additional analysis of the resultant embedding spaces. Contrastively trained models outperform other approaches on the classification tasks, delivering strong performance from limited labeled data and with fewer model parameter updates required. While metadata-based pre-training does not further improve classifications across the datasets, it yields interesting embedding cluster separability. All domain adapted LLMs outperform their publicly available general base LLM, validating the importance of domain-specialization. This research illustrates efficient approaches to instill healthcare competency in compact LLMs even under tight computational budgets, an essential capability for responsible and sustainable deployment in local healthcare settings. We provide pre-training guidelines for specialized healthcare LLMs, motivate continued inquiry into contrastive objectives, and demonstrates adaptation techniques to align small LLMs with privacy-sensitive medical tasks.    
    
\end{abstract}

\end{frontmatter}

\section{Introduction}

Large Language Models\footnote{LLMs refers to all relevant Pre-trained Language Models (PLMs) across all scales} (LLMs) such as the Bidirectional Encoder Representation Transformer (BERT) models based on masked language modelling (MLM), and Generative Pretrained Transformers (GPT) models based on autoregressive language modelling, have changed the research landscape entirely \cite{devlin-etal-2019-bert, wolf-etal-2020-transformers, attention-all-you-need}, and are generally seen as offering state of the art results on a number of popular Natural Language Processing (NLP) benchmark datasets and tasks \cite{devlin-etal-2019-bert, attention-all-you-need, Lester2021-power-scale-prompt, muennighoff_generative_2024}. The type of task different LLMs excel in generally depends on the architecture and pre-training objective: BERT-like LLMs, typically perform very well at embedding focused tasks like text classification, clustering and retrieval \cite{muennighoff_generative_2024}. Generative LLMs excel in generative tasks and form foundation for many chat-based APIs dominating the artificial intelligence (AI) landscape.

One common problem across LLMs, both embedding and generative focused LLMs, is a drop in performance on text data or NLP tasks from a specific domain (e.g. healthcare) to the one that the model was originally trained on \cite{huang2019clinicalbert, alsentzer-etal-2019-publicly, gpt-3-poor-few-shot-biomed, tunstall_efficient_2022, gutierrez_thinking_2022, sun_text_2023, tang_does_2023}. Aligning open LLMs with new domains and tasks remains a considerable issue, in particular for private datasets which remain unseen by open LLMs, e.g. UK National Health Service (NHS) based clinical datasets. Firstly, the LLM needs to be fine-tuned for each specific domain and task, and secondly, the LLM will suffer catastrophic forgetting, where it loses performance on other data and tasks in which it was originally trained. An intermediate approach to avoid these problems is to pre-train language models to producing good text embeddings that align well with possible downstream tasks to reduce the need for further fine-tuning, and thus improve training efficiency. Whilst certain approaches, such as Retrieval Augmented Generation (RAG), are becoming popular in attempting to imbue generally trained LLMs with external knowledge sources through embeddings, and thus reduce the need for domain adaption, the ability of the underlying LLMs involved is still often reliant on its domain pre-training. Moreover, the scale of the state of the art generative LLMs used today has become prohibitively large, whereas embedding focused LLMs remain much smaller, cheaper and efficient for many traditional NLP tasks. 
 
 In this paper, we explore the efficacy of altering the pre-training of smaller, BERT-like LLMs to align with the healthcare\footnote{we used the term healthcare to cover the more general NHS patient safety reports dataset used in this work that is not strictly an EHR} domain and downstream tasks of interest. The utility of small and resource-efficient LLMs for the healthcare domain is especially attractive where budget for compute may be limited and training on private data is required. Linked to this problem is the inability to utilise many cloud-based APIs with private data due to data governance laws and issues of confidentiality and security.

\subsection{Document-level label-free embeddings}
Standard language modelling objectives (e.g. autoregressive, masked language modelling) work on a word or token level, with the aim of representing words based on the context they appear in. For example, the LLM objective is operating on a token level loss objective, and embeddings are produced per token of the input sequence. Often the resultant token level embeddings appear performant for the task at hand, but many downstream tasks do not operate on the token level, but rather the sentence or entire document level e.g. document sentiment analysis, information retrieval, and similarity matching: of which LLMs are consistently state of the art. From a BERT-like LLM, a document level representation can be achieved in a number of ways based on the token level embeddings produced, with a common method being mean pooling i.e. the average of all token embedding representations of the last transformer layer \cite{devlin-etal-2019-bert, roberta, reimers_sentence-bert_2019}.

A potential issue with this relatively crude approximation of a document level embedding is that the Language Model (LM) pre-training objective did not encourage these representations to be useful or have a clear relationship to the loss function directly. A number of approaches have been explored to better train these LMs to produce sentence level embeddings through alignment tasks \cite{reimers_sentence-bert_2019, giorgi_declutr_2021, carlsson2021semantic}. It is common to utilise contrastive learning loss functions during pre-training to cluster embeddings from sentences or documents that are of the same type or even from the same global document closer together than those that are not. These approaches now incorporate a loss that is directly operating on the sentence or document level embedding and subsequently encourages the model to produce token embeddings that influence the resultant averaged embedding and embedding space. A potential difficulty for contrastive learning is deriving class labels that allow the creation of samples of positive and negative pairs that most contrastive loss functions require. A common approach relies upon a dataset's known labels, such as Natural Language Inference (NLI) datasets used in the training of sentence transformers\cite{reimers_sentence-bert_2019}. 

However, large labelled healthcare text datasets are rare and for the datasets explored in this paper we do not have extensive or completely labelled data. For this reason, we focus on methods that can use readily available structured meta data or use an unsupervised training regime, so no extra label annotations are required.

\subsection{The Healthcare Text Domain}


There are two related, but distinct, types of text that are often merged as a single category in the machine learning literature. These are "clinical text", defined as this one written by clinical professionals in Electronic Health Records (EHRs); and "biomedical text", this one written by researchers in scientific papers or books. The clinical domain, as opposed to the biomedical domain, is particularly difficult for general purpose LLMs. This is often thought to be due to the complexity and idiosyncrasies of the language used in clinical practice that inevitably reflects regional- or specialty-specific nomenclature. For instance, abbreviations are over-prevalent and often non-intuitive (e.g. using \texttt{"\#"} for broken bone, "bds" for twice a day, and many others \cite{nhs_abbreviations_2021}), and grammar and syntax is used sparingly in favour of faster writing. In contrast, the biomedical research literature tends to enforce a more consistent and agreed upon vocabulary, text follows a formal style, and descriptions and arguments try to be complete and analytic. 

There are several techniques to mitigate the problems that arise from merging these two domains, and these solutions typically rely on a form of transfer learning or domain adaptation. One typical approach is to continue training the LLM in the target specialist domain using the same language modelling objective, to better prepare the model for deployment in the new domain, and has delivered promising results for US based clinical datasets \cite{huang2019clinicalbert, alsentzer-etal-2019-publicly, taylor_clinical_2023}.

In this paper, we sought to compare and contrast the creation of embedding models in different healthcare datasets and in particular, two UK NHS datasets, and compare with general LLM alternatives. 

\subsubsection{The UK NHS}

Even in well-constrained use cases, such as pharmaco-vigilance for medication adverse events, clinical language patterns, idioms and idiosyncrasies in Electronic Health Record (EHR) data are notoriously difficult to work with \cite{luo_national_2020}.
Similarly, consultation with clinical colleagues in the NHS suggests that the routine clinical language recorded in EHR systems in the UK differs substantially from that used in large open-source medical text datasets. For instance, "A\&E" is sometimes used in the UK to denote the emergency department, while "ED" is used in the US and is more common in open-sourced datasets. In addition, the text of open-source medical datasets appears to consist of the milder examples that are closer to normal writing, and does not use abbreviations and grammatical transgressions as seen in the much more specialist text seen in actual EHRs. This can severely limit the deployment of current popular medical LLMs\cite{huang2019clinicalbert, chen_meditron-70b_2023}, which are often trained on these datasets. Whilst numerous biomedical or clinically trained LLMs exist, no publicly available LLMs exist for UK-based ``NHS language'' and most use data from the United States.


In this paper we investigate the influence that different LM pre-training objectives have on the performance of embeddings in downstream tasks. To make this investigation systematic, we examined three pre-training methods across three healthcare datasets, and inspected the results using dataset-specific sequence classification tasks,  embedding distance metrics and qualitative analyses. Overall, we aim to ascertain what benefits different pre-training methods provide when re-using general-domain LLMs to the clinical domain, and in particular to NHS datasets, such that deployment becomes more amenable than in prior work. 

\subsection{Motivation and Related Work}
Improving the embeddings representation of LLMs in the general domain has been studied extensively and influences this work directly: notably the Deep Contrastive Learning for Unsupervised Textual Representations (DeCLUTR) paper \cite{giorgi_declutr_2021}, sentence transformers \cite{reimers_sentence-bert_2019} and SimCSE \cite{gao_simcse_2022} works show great promise in improving BERT-style LLMs representation of sentences and documents through contrastive methods.
Similar works investigating the augmentation and changing of the LLM pre-training objectives with external knowledge-graphs include BioLinkBERT \cite{yasunaga2022linkbert} and Dragon \cite{yasunaga_deep_2022}. SAPBert sought to improve named-entity-recognition and linkage through alignment of embeddings for entity synonyms \cite{liu2021self-SAP-bert}.

Recently, there has also been promising work showing how task specific contrastive loss functions combined with sentence transformers can adapt LLMs to downstream tasks with relatively few training samples \cite{tunstall_efficient_2022}. Work has also sought to enhance the ability for generative LLMs to produce both good embeddings and generation using a combination of instruction tuning and embedding loss functions \cite{muennighoff_generative_2024}, however this still relied on comparatively larger models (e.g. those with more than 7 billion parameters).

The key contributions of this work are the introduction of a note category pre-training objective, the development of several LLMs for different NHS datasets, and the exploration of resource constrained pre-training and downstream adaptation.

 %

\section{Methods}

\subsection{Datasets and downstream tasks}\label{datasets}

As is typical, to explore the effect of different pre-training methods on the LLMs we sought to evaluate them on a set of dataset specific downstream tasks. The focus was primarily on the differences in document-wide embeddings dependent on the pre-training received. Thus we opted to derive document classification tasks using the categorical variables available for each of the datasets, with a focus on potential clinical use-cases. The specific tasks available for each dataset are outlined below, with an overview of the data distributions and their respective downstream task is provided in Table \ref{table:pseudo-dataset-description}. For brevity, all the downstream tasks are document classification tasks whereby a document is to be assigned a particular class label.

\subsubsection{MIMIC-III}
The first dataset we used is the Medical Information Mart for Intensive Care III (MIMIC-III) \cite{Johnson2016}, a medical dataset developed by the MIT Lab for Computational Physiology. It is comprised of de-identified EHR records associated with 38,597 critical care patients and 58,976 intensive care unit (ICU) admissions at the Beth Israel Deaconess Medical Center between 2001 and 2012. Data includes demographics, vital signs, laboratory tests, medications, caregiver notes, imaging reports, and mortality in and out of hospital. While the clinical tasks presented here may benefit from utilising the multi-modal data available for each patient, we focus on the use of free text clinical notes.


\subsubsection*{ICD-9 Triage (M-Tri)}
We utilise the ICD-9 codes associated with discharge summaries in the MIMIC-III dataset to derive a \textit{triage} classification task, originally published in our previous work \cite{taylor-clinical-prompt-ieee}. The motivation for this task is to represent a realistic use-case within a hospital setting, whereby patients admitted to an ICU will be treated and then ``stepped down'' (discharged) to another ward or team to continue treatment when they no longer require an ICU. The result is a mapping between particular ICD-9 diagnosis codes and a corresponding \textit{destination} department, of which we derive the following seven post-ICU destination teams: \textit{Cardiology}, \textit{Obstetrics}, \textit{Respiratory Medicine}, \textit{Neurology}, \textit{Gastroenterology}, \textit{Acute or Internal Medicine}, and \textit{Oncology}. For further details, see original implementation details in \cite{taylor-clinical-prompt-ieee}.

\subsubsection{Oxford Health Foundation Trust - OHFT}
The second dataset is from the Oxford Health NHS Foundation Trust (OHFT), a regional UK-based provider of specialist mental healthcare covering Oxfordshire and Buckinghamshire. The dataset contains full historical EHR data for approximately 200,000 patients spanning over a decade, and within this access to around 8 million de-identified clinical notes.


\subsubsection*{Triage Team Association (O-Tri)}

In the UK, mental health services are structured into primary, secondary, and tertiary levels. Most patients (96\%) needing specialist care are referred to and treated by community mental health teams (CMHTs) \cite{NHS-dig-2020}. Referrals contain information written by the referring doctor or professional and are triaged by the receiving CMHT into: a) accept to the team for assessment, b) reject due to insufficient information, or c) route to a sub-specialty team if warranted.

Structured EHR data on referral and discharge dates establishes which team accepted the patient, though local administrative variations required developing supplementary heuristics in collaboration with OHFT clinicians. We created a classification task to identify the accepting triage team for a given referral from the subset of accepted referrals. Specifically, given a random clinical note, the task is to determine which referral team it likely belongs to based solely on its free text contents.

\subsubsection{NHS Patient Safety Incident Reports - PSIR}
The third dataset is a large, national collection of free-text documents relating to all types of patient safety incidents in the NHS, from the National Reporting and Learning System (NRLS). More information about the dataset and collection can be found on the official website \cite{noauthor_nhs_nodate}. We worked with a sub-sample of approximately 2.3 million de-identified reports produced in the financial year 2019/2020, with the goal of supporting more efficient analysis of the dataset within NHS England, and integration of the learning into the newer Learn from Patient Safety Events (LFPSE) service.

\subsubsection*{IN05 Level 1 - Incident Category (P-Cat)}
The labels of incident category at the first level are a standard field provided alongside each incident report which is used to detail the type of the incident.  There are 13 classes present in the dataset, and we have chosen not to include a second level of labels which is included in the wider NRLS dataset.

\subsubsection*{PD09 - Incident Degree of Harm  (P-Sev)}
The degree of incident severity categories are an ordinal scale ranging from no harm (1) to death (5) collected for all incidents.  We simplify a pseudo task related to incidence severity prediction in the following way: the 1-5 incident severity labels given with the free text reports are skewed, with the vast majority being attributed a 1 (or no harm).  We create a much smaller and balanced binary classification dataset which bins incidence labels into low (severity categories 1-3) and high (severity categories 4-5) severity. 



\subsubsection{Note Category - All Datasets}
\label{subsubsection:notecat}
The various origins and purposes of these clinical notes are partially captured by their \textit{note category} assignment (or associated metadata). These categories can be be seen as a form of structured knowledge pertaining to the organisation of the part of the healthcare system of focus. This is a commonly captured field in healthcare datasets and while the exact use and meaning of this may differ between the respective datasets, broadly each dataset contains this field to record the professional role of the person who produced the note or the departmental origin. We expect that note category may be a valuable signal for the pre-training itself because, for example:
\begin{itemize}
    \item a note entered by a social worker will contain terminology and describe concepts relevant to a patient's social circumstances
    \item a note entered by an occupational therapist will emphasise  functioning and the patient's capacity for everyday tasks
    \item a note entered by a physician will emphasise clinical state, examinations and treatment plans
    \item a note entered by a care coordinator will reflect progress on executing a management/care plan for the patient
\end{itemize}

Therefore, embeddings arising from the contents of documents with different note categories should capture the vocabulary, concepts and semantics of information routinely recorded by different healthcare professionals and their intended use. In each of our dataset we identify a \textit{note category}: for both \textit{MIMIC-III} and \textit{OHFT}, the note category reflects the clinical purpose of the document that relates to the profession of the individual making the note e.g. \textit{Nursing}, \textit{Doctor}, \textit{Social worker}. Whereas for \textit{PSIR}, the category variable (RP02 on the database) reflects the care setting in which the patient safety incident occurred such as: Acute / General Hospital, various Community groupings, and General Practice. 


\subsection{Data splits}

For each of our datasets there are a substantial number of available clinical documents, even in the smallest there are over 2 million individual samples. To facilitate the development of multiple models and experiments within a resource restricted setting, we developed and trained initial pipelines at various smaller scales. We then chose to train and test models on a maximum sub-sample of 250,000 documents as this would be large enough to show trends in the data, whilst keeping resource utilisation to a manageable level. Furthermore, where possible we utilised the unique patient identifiers to ensure no individuals data was both in the training and evaluation sets. Future work could seek to train and evaluate with a larger amount of data and focus on the magnitude of differences in results found. 


\subsection*{Length of documents}
An important feature of a text dataset is the number of words, or tokens, contained in each individual sample (document). Transformer based language models such as RoBERTa\cite{liu_roberta_2019} can only handle a maximum of 512 tokens on even the most modern GPUs, due to the complexity of the self-attention calculations which increases exponentially with the number of tokens \cite{longformer2020}. The average document length, as well as the distribution of lengths, varies considerably in our three datasets, although a large portion fit within the maximum sequence length for our chosen models. As is standard, documents that exceed the maximum token length are truncated. Whilst numerous approaches have been developed to mitigate this problem, such as introducing local sliding windows \cite{longformer2020}, key-value caching \cite{pope_efficiently_2022}, and flash attention \cite{dao_flashattention_2022}, we opt to focus only on the standard transformer attention used by RoBERTa.  We thus limit the context window to 512 tokens, which we see as sufficient for the investigations presented in this work. For more details of the document lengths for each respective dataset, see \ref{appendix:dataset-details}.



\begin{table*}[h]
\caption{Downstream classification dataset statistics. Task names are also given short-hand codes used throughout the rest of the paper to better fit tables and figures.*P-Cat is not the note category used in pre-training for the PSIR dataset}
\centering
\footnotesize
    \begin{tabular}{c c c c c}
    \addlinespace
    \toprule
   \textbf{ Dataset} & \textbf{Task (Acronym)} & \textbf{\# labels} & \textbf{\# train samples} &\textbf{ \# test samples} \\ 
    \midrule
    Mimic-III & Note category (M-Cat) & 8 & 1,600 & 4,000 \\
    \midrule
    Mimic-III & ICD-9 Triage (M-Tri) & 7 & 1,400 & 3,150\\
    \midrule
    OHFT & Note category (O-Cat) & 10 & 10,000 & 2,500 \\
    \midrule
    OHFT & Referral team relation (O-Tri) & 5 & 6,250 & 2,500 \\
    \midrule
    PSIR & IN05: Category (P-Cat) & 13 & 26,000 & 2,600 \\
    \midrule
    PSIR & PD09: Severity (P-Sev) &  2 & 14,000 & 14,000 \\
     \bottomrule
    \end{tabular}
    \label{table:pseudo-dataset-description}
\end{table*}

\subsection{Language modelling - Preliminaries}


The pre-training of a standard LM involves a corpus of $D$ text documents $X=\{X_1, X_2, ..., X_D\}$ and two functions, $f_{enc}$ and $f_{head}$. Each document $d$ consists of a sequence of $T$ tokens $X_d = (x_1, x_2, x_3...x_T)$ which is first passed through $f_{enc}$ to produce a contextualised vector for each token, $H_d$ = $(h_1,h_2,h_3...h_T)$. $f_{head}$ then uses $H_d$ for whichever self-supervised pre-training task chosen and to perform subsequent downstream tasks during fine-tuning. Because $T$ may vary from sample $d$ to sample $d'$, to represent the whole sequence of $T$ tokens as a fixed-length vector, as is required for many tasks, we used a pooling function $g$ to take the mean of all token level embeddings of the sample, $H_d=(h_1, h_2, ..., h_T)$, which has proven a reasonable approach to represent a sequence of text \cite{giorgi_declutr_2021, reimers_sentence-bert_2019}. Accordingly, whole sequence embeddings $e_d = g(f_{enc}(H_d))$.

\subsubsection{Continued Masked Language Modelling}

The first pre-training method is the standard formula for MLM, a commonly used objective for pre-training language models that randomly replaces or \textit{masks} a proportion of tokens of the input with a special $[MASK]$ token. This essentially corrupts the original input and the objective of the model is to predict which tokens should appear in the \textit{masked} positions, a form of gap filling. 



The standard MLM loss function is given as follows:


\begin{equation} \label{eq:mlm_loss}
    \mathcal{L}_{mlm}(X , Y) = -\sum^{N_c}_{n=1}W_{n}(\sum^{|V|}_{i=1} Y^{n}_{i}\ln(f_{lm}(X)_{i}^{n}))
\end{equation}
where $X$ is the input of the model, $Y$ denotes MLM labels which is a collection of $N_c$ one-hot vectors each with the size of $|V|$ where $|V|$ is the size of the vocabulary of the model and $N$ is the number of input tokens\footnote{Note that one-hot vectors for non-masked tokens are zero vectors.} and $W_{n}$ is $1$ for masked tokens and $0$ for others. This ensures that only masked tokens will contribute to the computation of loss. $f_{lm}$ represents the encoder model with a language modelling head whose output is a probability distribution vector with the size of the vocabulary ($|V|$) for each token.

\subsubsection{Contrastive Loss Pre-training}

A common approach to improving separation of classes in an embedding space utilises contrastive learning, which uses a loss function that aims to encourage semantically close or same-class members whilst pushing apart other-class members \cite{gunel_supervised_2021}.
It assumes a set of paired examples $\mathcal{P} = \{(X_i, X^{+}_j) \}_{i,j=1}^P$, where $P$ are the number of samples, and documents $X_i$ and $X_j$ are semantically related documents. The derivation of \textit{paired} examples is the important, and arguably most difficult part when well-defined labels are not present. We therefore first opt to utilise methods that are unsupervised or self-supervised:

\subsubsection*{DeCLUTR}

As we want to generate a model that can produce good document level embeddings, we also explored an self-supervised cluster alignment technique used to produce the DeCLUTR model \cite{giorgi_declutr_2021}. DeCLUTR stands for Deep Contrastive Learning for Unsupervised Textual Representations, and uses a contrastive loss function to encourage sentence or document level embeddings that are taken from the same document type or class to be closer together in the learned embedding space.  DeCLUTR utilises a self-supervised contrastive loss function called InfoNCE (Noise-Contrastive Estimation) which aims to identify positive pairs in a set of samples also containing numerous negatives. 

\begin{figure}[!htb]
    \centering
    \includegraphics[scale=0.95]{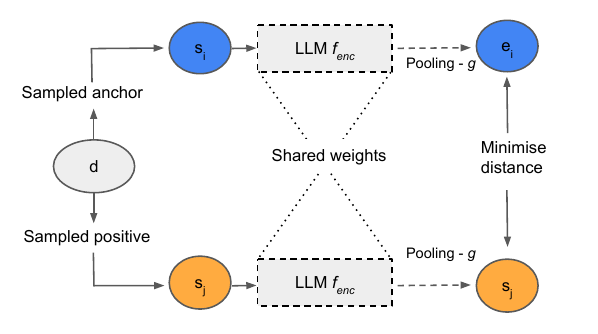}
    \caption{Adapted from \cite{giorgi_declutr_2021}. Overview of the DeCLUTR training process. We sample anchor spans $s_i$ and positive spans $s_j$ from each document $d$ in a minibatch of size $N$. For simplicity, we show $A=P=1$, where $A$ and $P$ are the number of anchors and positives per document. The spans are encoded by $f_{enc}()$ and pooled by $g(\cdot)$ to get embeddings $e_i = g(f(s_i))$ and $e_j = g(f(s_j))$. The encoder and pooler are trained to minimize the distance between positive span pairs while maximizing the distance to negatives (omitted for simplicity). }
    \label{fig:declutr}
\end{figure}

 During training, a batch of $P$ anchor-positive span pairs is taken from document $d$. Each of the spans are separately encoded. The anchor and positive embeddings, $s_i$ and $s_j$, are then compared for their cosine similarity by taking their dot product. The objective of the training procedure is to maximise the cosine similarity of the $P$ matching anchor-positive pairs and minimise that of the remaining $P^2 - P$ non-matching pairs. For a given batch, the cosine similarities are then used to calculate the probability that a given pair is a match, which can be defined as: 

\begin{equation}
    P(s_i, s_j; \tau) = \dfrac{exp(s_i \cdot s_j / \tau)}{\sum_{k\ne i,j} exp(s_i \cdot z_k / \tau)} 
    \label{eq:match_prob}
\end{equation}

\noindent
where $\tau$ is a trainable temperature parameter. This results in the InfoNCE loss which symmetrically measures the success in maximising the similarity of matches and minimising the similarity of non-matches, defined as: 


\begin{equation}
\begin{split}
    L_{\text{InfoNCE}} &= -\dfrac{1}{2} \Biggl[ \dfrac{1}{P} \sum^P_{i,j=0} \log P(s_i, s_j; \tau) + \dfrac{1}{P} \sum^P_{i,j=0} \log P(s_j, s_i; \tau) \Biggr]
\label{equation:infonce}
\end{split}
\end{equation}

\paragraph{DeCLUTR Data Sampling}
For the DeCLUTR models, we initially only manipulated the sampling strategy, through changing the maximum length of documents spans to be used for the creation of anchor-positive pairs, in our experiments. We kept the actual sampling methodology the same. 
 As per the original paper, the minimum number of tokens for a document to be sampled is: $2 \times A \times S_{max}$ where $A$ is the number of anchors to be sampled, and $S_{max}$ is the maximum span length. For example, the minimum span length for a document with 2 anchors and a maximum span length of 512 would be 2048. 

The span length is thus a key parameter when training using the DeCLUTR regime and we have not been able to exhaustively explore all possible combinations. We instead opted for a span length that aligned with our average document length in each dataset. The sampling strategy for generating anchors and positive pairs was also fixed to adjacent across all datasets. We recognise that future work could seek to explore this space further. Examples of the sample distributions based on the minimum document length for each dataset are presented in the table in  \ref{appendix:declutr-sampling}.


\subsubsection*{Note Category as a pre-training signal}

As a third pre-training method, we utilised a known categorical variable which appears in some form across all three datasets, the ``note category'' of the document as described in Section \ref{subsubsection:notecat}.

We formulate this task as a replacement of the original \textit{next sentence prediction} task used in BERTs implementation \cite{devlin-etal-2019-bert}. Whole sequence embeddings, $e$, are  fed to a classification head $f_{head} (\cdot)$, which has the task of calculating the logits $y_j$ of each of the possible $c$ classes $j \in \textbf{C}$. A softmax operation $\sigma$ is applied to the logits to produce a normalized probability score that $\boldsymbol{x}$ belongs to each of $c$ possible classes. For one sample with the vector representation $\textbf{e}$, the probability of the sample belonging to class $j$ is:

\begin{align*}
    f_{head}( \textbf{e} ) &= \left[ y_1, y_2, ..., y_c \right], \\
    P( j ) &= \sigma( \left[ y_1, y_2, ..., y_c \right] ) = \frac{ exp( y_j ) } { \sum_{k=1}^{c}{ exp( y_k )} }
\end{align*}
where  $f_{head}(\cdot) : \mathbb{R}^{m} \xrightarrow{} \mathbb{R}^c$, and $y_j \in \mathbb{R}$ for all $j$.  The classification head can have any number of layers (depth) $d \in \mathbb{N}$, but here we have opted to set $d = 2$ throughout. 

Following this, the loss for the note category pre-training can be defined by the standard cross-entropy formula:

\begin{equation}
    \mathcal{L}_{note} = -\sum_{c=1}^N y_c \log(p_c)
\end{equation}


\begin{figure}[htp]
    \centering
\includegraphics[width=1\textwidth]{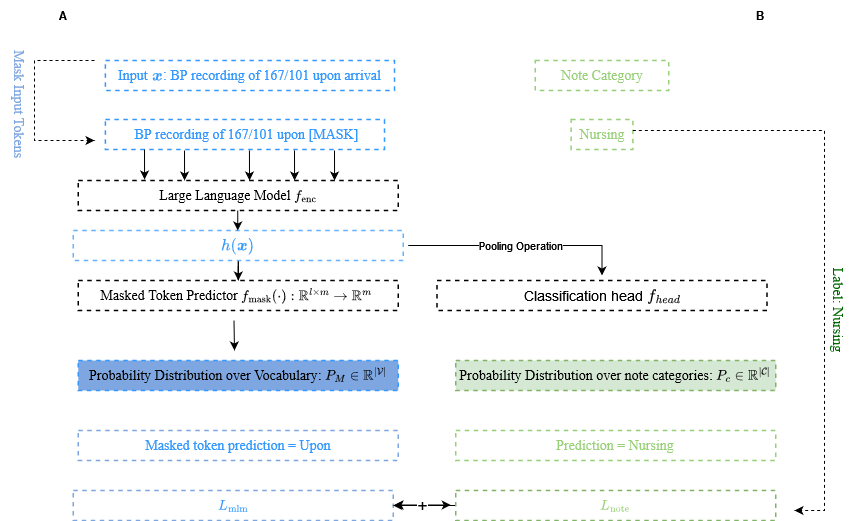}
    \caption{Overview of our note category pre-training approach. On the left side (\textbf{A)} shows the flow of the input sequence ($\textbf{x}$) through the standard MLM pipeline, and on the right side (\textbf{B}) shows the integration of the associated note category label in parallel. The MLM and note category classification objectives are jointly optimised with each document.}
    \label{fig:note-category-diagram}
\end{figure}

With both contrastive pre-training objectives outlined, we combine them with standard MLM to form a joint loss function. To discourage the pre-training objective to over-represent the note category classification task, we also applied an optional weighting $w$ to the loss, as shown in Equation \ref{eq:mlm_contrastive}, where $L_{\text{contrastive}}$ is either $L_{\text{InfoNCE}}$ (for DeCLUTR) or $L_{\text{note}}$ (for note contrastive).

\begin{equation}
    L = L_{\text{MLM}} + w(L_{\text{contrastive}})
    \label{eq:mlm_contrastive}
\end{equation}




\subsection{Adapting LLMs for downstream classification Tasks}

In order to use the various further pre-trained LLMs for downstream classification tasks, we used the traditional fine-tuning approach. Conventional fine-tuning can be achieved by adding task-specific layer(s) or an entire multi-layer perceptron (MLPs) to the LLM. The exact approach to processing the LLM output is dependent on the task.

In our use case of sequence or document classification, the downstream task head is an MLP $f_\text{MLP}(\cdot)$ made of up 2 linear layers which takes the pooled sentence embedding output by the LLM, $e$, as input and generates an $n\text{-dimensional}$ vector, where $n$ is the number of classes. A softmax operation is applied to the resultant output vector in order to generate probabilities of each class: 


$$ P(y \mid \boldsymbol{x}) = \frac{\exp{ ((f_\text{MLP}\left(h(\boldsymbol{x})\right)_y)}}{\exp{\left( \sum_{i=1}^{n} f_\text{MLP}(h(\boldsymbol{x}))_i \right)}}.$$ Since the additional MLP block and LLMs are modular, their respective parameters are stored separately, and we can opt to freeze the parameters of one or the other.



\subsection{Efficiency gains from LLM freezing and few-shot training}   
 We sought to explore the potential of using the embeddings produced by the different LLMs without any further fine-tuning in relation to a given downstream task, keeping the LLM body frozen, or by freezing different numbers of layers of the LLM model. Moreover, we explored the performance of the LLMs on the downstream tasks by fine-tuning with different numbers of training samples per class in a few-shot training setup, similar to previous works \cite{taylor_clinical_2023}. 

\subsection{Document Embeddings Analysis}
Beyond downstream classification performance, we chose a selection of different metrics of the LLMs embedding space to discern any clear differences produced by the varied pre-training objectives used.
\subsubsection{Uniformity and Alignment}
We follow a similar analysis plan to the authors of the SimCSE paper \cite{gao_simcse_2022} to probe the \textit{quality} of the embedding space through measures of \emph{alignment} between in-class pairs and \emph{uniformity} across the entire space. Alignment calculates expected distances between the paired instances, which in our case was embeddings of documents within the same class (a proxy for positive pairs).

\begin{equation}
\ell_{\text{align}}\triangleq \underset{(x, x^+)\sim p_{\text{pos}}}{\mathbb{E}} \Vert f(x) - f(x^+) \Vert^2
\end{equation}

Conversely, {uniformity} measures how well the embeddings for all documents, regardless of class, are uniformly distributed. Together these metrics help illuminate how the embedding spaces represent within class samples, which should remain close together and random, unrelated samples should be scattered.
    

\begin{equation}
    \label{eq:uniformity}
    \ell_{\text{uniform}}\triangleq\log \underset{~~~x, y\stackrel{i.i.d.}{\sim} p_{\text{data}}}{\mathbb{E}}   e^{-2\Vert f(x)-f(y) \Vert^2} 
\end{equation}

\subsubsection{Cosine Similarity and Clustering}
Given we produce a set of embeddings for our texts with known labels, we can run a number of analyses utilising the raw vector embeddings for each note, as well as dimension reduced embedding spaces. For our analysis we opted to look at simple cosine similarity within, and between known classes for each dataset and a simple graph network analysis. 

Graph network analysis provides another lens for understanding the structure of embedding spaces. In this approach, documents are represented as nodes in a graph, and we define our edges connecting notes that have a cosine similarity above a defined threshold. The connectivity of the graph and formation of connected subgraphs can reveal insights about how notes are positioned in embedding space.

\section{Implementation details}\label{implementation}

\subsection{LLM model setups}
We performed a large and varied number of experiments across the three datasets, with different training regimes and requirements. An overview of these is presented in Table \ref{table:model_setups}, although this is a curated selection intended to provide broad coverage of the models and training objectives used.

\subsection{Training overview}

A secondary objective of this work was to investigate the efficient pre-training of small LLMs given resource constrained environments (where large suites of GPUs are not available nor desirable). Further, Language modelling from scratch is often expensive and hardware dependent, thus we chose to continue the \textit{pre-training} phase extending already pre-trained general domain LLM. Importantly, we conducted training, evaluation, and hyperparameter exploration with the use of a single local GPU, similar to other works attempting to complete full pre-training with a single GPU given a fixed amount of time \cite{geiping_cramming_2022}. For more information of the hardware setups in each case, please see \ref{appendix:hardware}.

\begin{table*}[hpt]

\centering
\footnotesize

\begin{tabular}{c c c @{\hspace{0.5\tabcolsep}} c}

\toprule
\textbf{Clinical Dataset} & \textbf{Domain Pre-training} & \textbf{Model size} & \textbf{Model Name} \\
\midrule
None & None & 124.6 & RoBERTa-base \\
\midrule
Mimic-III & MLM & 124.6 & RoBERTa-mimic \\

Mimic-III & MLM + DeCLUTR & 124.6 & RoBERTa-mimic-DeCLUTR \\

Mimic-III & MLM + Note & 125.1 & RoBERTa-mimic-note \\

\midrule

OHFT & MLM & 124.6 & RoBERTa-OHFT \\

OHFT & MLM + DeCLUTR & 124.6 & RoBERTa-OHFT-DeCLUTR \\

OHFT & MLM + Note & 125.1 & RoBERTa-OHFT-note \\

\midrule

PSIR & MLM & 124.6 & RoBERTa-PSIR \\

PSIR & MLM + DeCLUTR & 124.6 & RoBERTa-PSIR-DeCLUTR \\

PSIR & MLM + Note & 125.1 & RoBERTa-PSIR-note \\

\bottomrule
\end{tabular}
\caption{Overview of the different datasets and LLM pre-training. Domain pre-training refers to whether this model has been explicitly pre-trained using the related clinical dataset. We will use this table to define the model names that will be used throughout the results section. Each model uses RoBERTa-base as the base model.}
\label{table:model_setups}
\end{table*}


\section{Results}

After extending the pre-training of the different LLMs using the methods outlined above (MLM, DeCLUTR, and Note Contrastive), we have 12 LLMs to compare, outlined in Table \ref{table:model_setups}. Given these LLMs now produce different representations of text, we evaluate them across each of the respective datasets and associated downstream tasks. We present two approaches to evaluating these embeddings: downstream classification performance, and embedding space analysis.

\subsection{Downstream Classification Performance}
We report results for using the LLMs in different settings.First, keeping the base LLM frozen during fine-tuning, and only update the weights of the classification head to assess the utility of the pre-trained embeddings with no further updates (effectively the LLMs here are pure feature extractors). We compare this with freezing varying numbers of LLM layers, and also fully fine-tuning the entire LLM on the downstream task. 

\subsubsection{Evaluation - all tasks}
The evaluation performance for each of the domain pre-training methods across the different datasets and tasks is presented in Table \ref{tab:max-f1-all-tasks}  (best over five epochs), with DeCLUTR models performing best when the LLM remains frozen during fine-tuning. In the full fine-tuned setting, MLM models generally perform marginally better. Across all tasks and fine-tune settings, the models with no domain pre-training achieve the lowest performance. For individual performance metrics for each model across each dataset and task, including results for alternative open-source clinical LLMs, please see \ref{appendix:downstream-performance}.


\begin{table*}
    \centering
    \footnotesize
    \begin{tabular}{lccccccc}
        \textbf{PLM} & \textbf{Domain pre-training}  & \textbf{M-Cat} & \textbf{M-Tri} & \textbf{O-Cat } & \textbf{O-Tri}  & \textbf{P-Cat} & \textbf{P-Sev}  \\
        \midrule
        Frozen      & None  & 0.766 & 0.314 & 0.319 & 0.423 & 0.384 & 0.65 \\
                    & MLM &\textbf{0.867} & 0.439 & 0.355 & 0.552 & 0.498 & 0.739 \\
                    & MLM + DeCLUTR &0.859 & \textbf{0.711} & \textbf{0.411} &\textbf{0.601} &\textbf{0.555} & \textbf{0.748} \\
                    & MLM + Note & - & 0.471& - &0.546 & 0.450 & 0.714 \\
        \midrule
        Finetuned& None &0.991 &0.827 & 0.593 &0.766  & 0.655  &0.837 \\
                & MLM &\textbf{0.991} &\textbf{0.846} &\textbf{0.629} &\textbf{0.779} &0.660 &0.842 \\
                & MLM + DeCLUTR &0.988 &0.844 &0.613 &0.765 &0.653 &0.844 \\
                & MLM + Note & - & 0.836 &- & 0.757  & \textbf{0.663} &\textbf{0.847} \\
    \end{tabular}
    \caption{$F1$ macro across all datasets and classification tasks based on domain pre-training received. We report the maximum $F1$ macro achieved over 5 epochs of training per model.}
    \label{tab:max-f1-all-tasks}
\end{table*}

\subsubsection{Few-shot sampling}
\begin{figure}[ht]
    \centering
\includegraphics[width=1.0\textwidth]{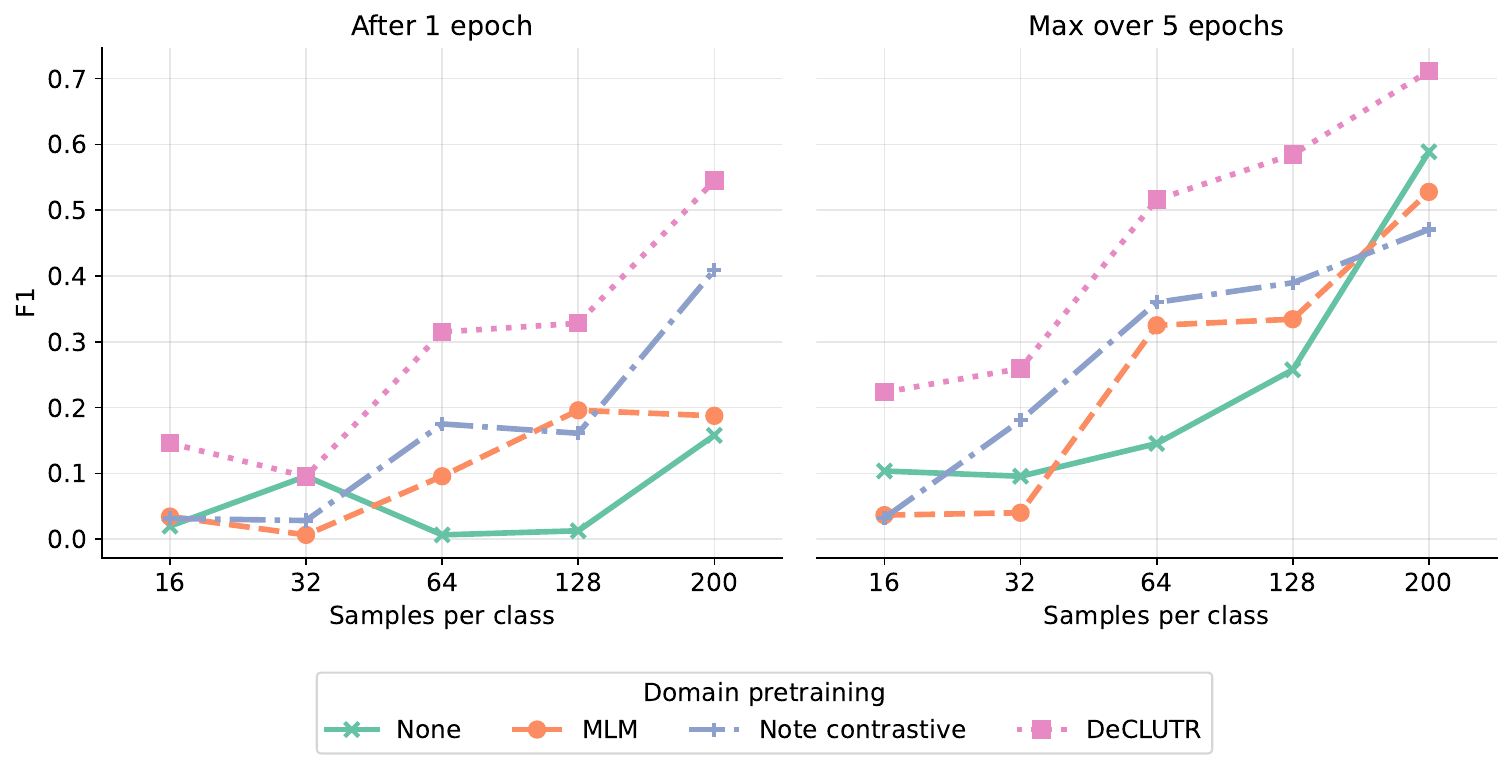}
    \caption{F1 macro score on evaluation set for the MIMIC-III ICD-9 Triage task with frozen LLMs trained with different sample sizes per class.}
    \label{fig:f1-frozen_triage}
\end{figure}

Additionally we focus on training with sub-samples of the training set, representing a setting where producing annotations is often a major limiting factor (e.g. due to lack of expert time or prohibitive expense) in Fig. \ref{fig:f1-frozen_triage}. For brevity we only include this analysis for the MIMIC-III ICD-9 Triage task, however similar patterns were seen across all datasets and tasks.

\subsubsection{Effect of freezing layers}
\begin{figure}[ht]
    \centering
    \includegraphics[width=1.0\textwidth]{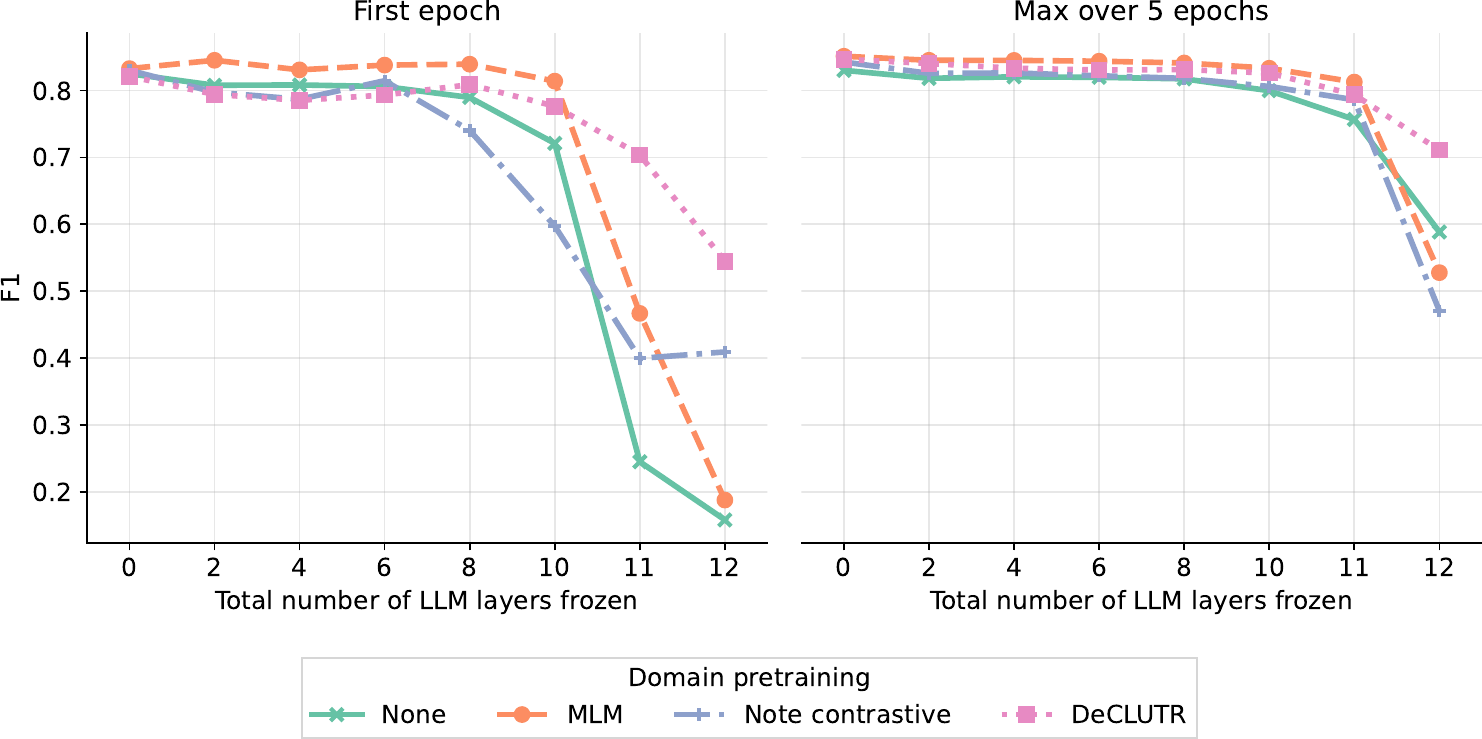}
    \caption{F1 macro score on evaluation set for the MIMIC-III ICD-9 Triage task with varying transformer layers frozen (all models utilised a 12-layer RoBERTa architecture).}
    \label{fig:ICD-9-triage-frozen-layers}
\end{figure}

To investigate the influence of freezing different layers of the LLM, we present results based on freezing an increasing number of consecutive layers of the LLM in Fig. \ref{fig:ICD-9-triage-frozen-layers}.


\subsection{Document Embeddings Analysis}

The main result of training a LLM is a model that has captured aspects of how human language is organised, encapsulated by its resultant embedding space. The pre-training objective has a direct impact on the embedding space, as well as the domain targeted downstream tasks. For any language modelling task or text-based downstream task, the word or sentence derived contextualised embeddings are the numerical representation of that knowledge obtained during pre-training. 

We have seen substantial differences in the usefulness of different LLMs embeddings for downstream tasks, especially when no fine-tuning occurs. To attempt to understand this aspect in more detail, we present a exploration of the embedding spaces of different LLMs for the MIMIC-III ICD-9 triage task. For more embedding analysis details, see \ref{appendix:cluster-analysis}.


\subsubsection{Cosine Similarity}

\begin{figure}[htp]
    \centering
    \includegraphics[scale=0.6, ]{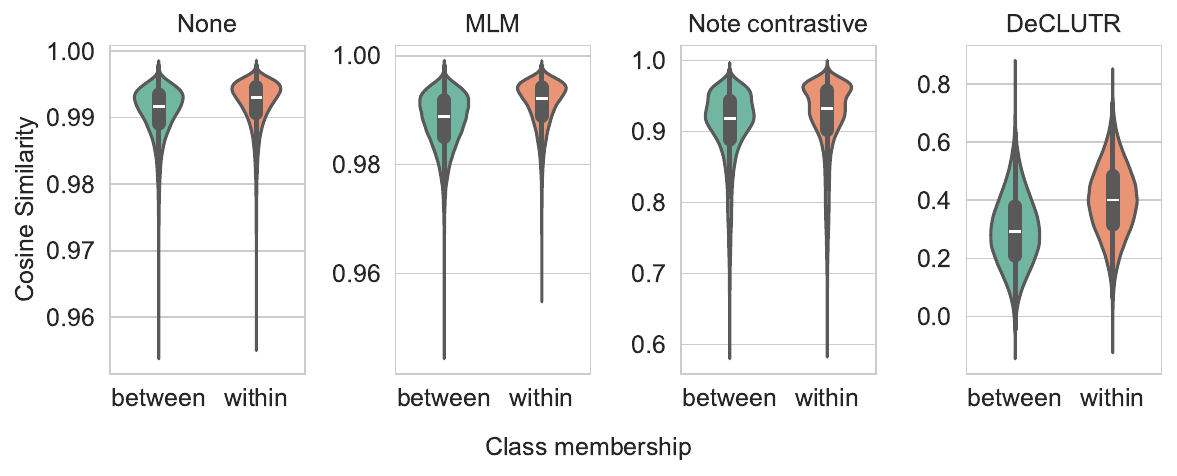}
    \caption{Cosine similarity of document embeddings within and between classes for the MIMIC-III ICD-9 triage dataset. Note the y-axis scales are separate for each subplot, this is due to the large differences in value ranges between models.}
    \label{fig:embeddings-cosine}
\end{figure}

A common approach to measuring the characteristics of an embedding space, especially when classes are known, is to look at the distances between embeddings of each class within the embedding space. As expected, the LLMs which utilised a contrastive loss function (RoBERTa-mimic-DeCLUTR and RoBERTa-mimic-note) exhibit a much greater separation of embeddings, with a much wider range of cosine similarity values. However, the differences between and within class members shows a very similar pattern amongst all models, see Fig. \ref{fig:embeddings-cosine}. 

\subsubsection{Alignment and uniformity}

\begin{figure}[htp]
    \centering
    \includegraphics[scale=0.5,]{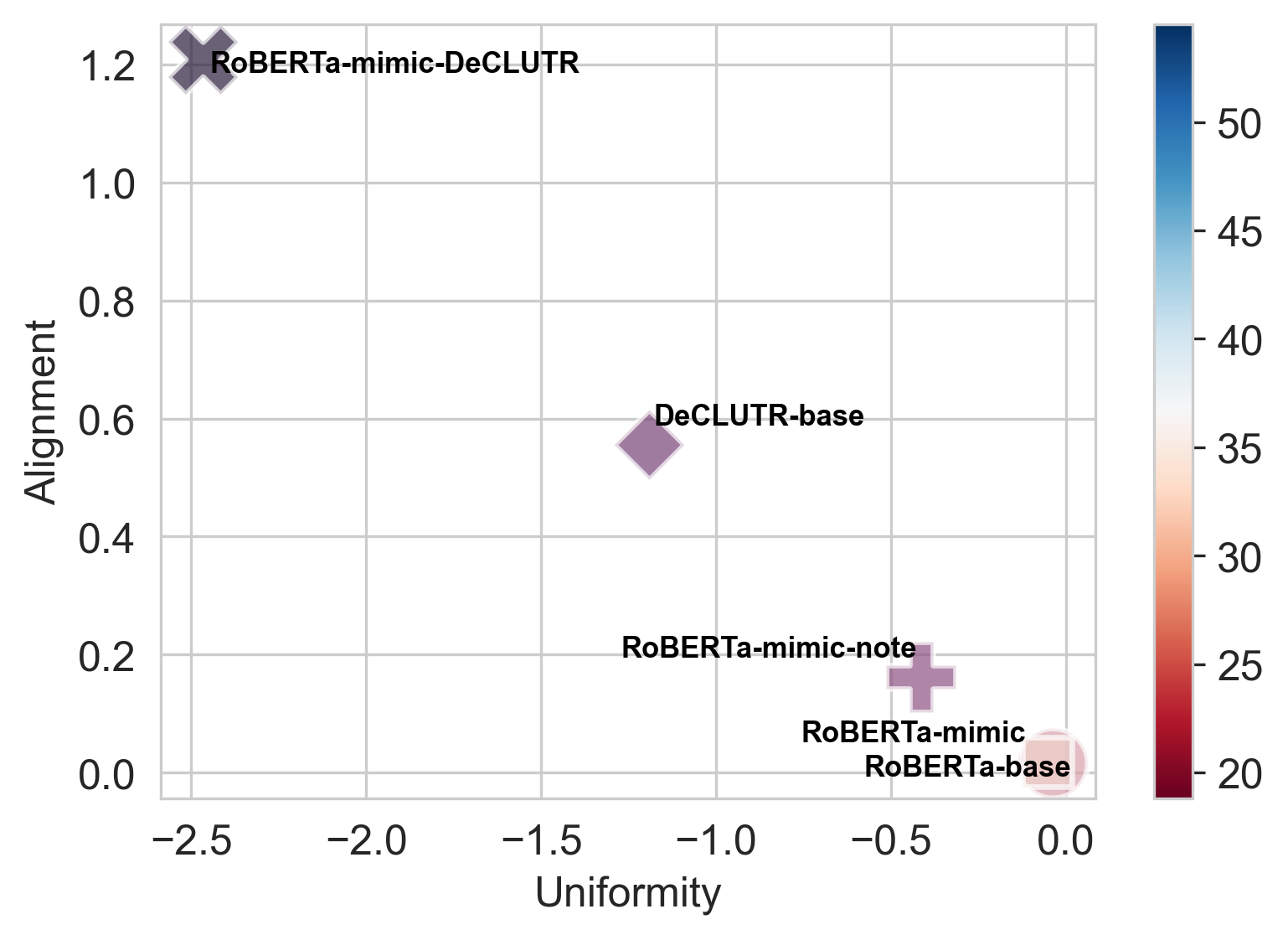}
    \caption{Uniformity vs. alignment for the different LLM setups for the MIMIC-III ICD-9 task embeddings. The colorbar represents the corresponding F1 score on the ICD-9 triage classification task using these frozen LLM embeddings. }
    \label{fig:uniform-align}
\end{figure}

An example of comparing uniformity versus alignment is presented in Fig. \ref{fig:uniform-align}, which highlights large differences between the LLMs dependent on their pre-training objective. Most notably, the DeCLUTR models appear to have produced an embedding space with a high diversity and hence low uniformity amongst all classes, but with a high alignment score, implying within class embeddings remain relatively far apart.

\subsubsection{Network analysis}

Results of a simple graph network analysis with varying cosine similarity thresholds is provided in Table \ref{tab:network-analysis}. 

\begin{table*}[htp]

\footnotesize

\centering

\begin{tabular}{llrrr}

\toprule
\textbf{Model name} & \textbf{Cos. threshold} & \textbf{\# Components} & \textbf{Avg. degree} \\

\midrule
RoBERTa-base & 0.995 & 1 & 317.896  \\
 & 0.996 & 4 & 54.355 \\
 & 0.997 & 8 & 24.128 \\
 & 0.997 & 42 & 46.628 \\

\midrule
RoBERTa-mimic & 0.994 & 2 & 160.150 \\
 & 0.995 & 7 & 168.512 \\
 & 0.996 & 16 & 82.579 \\
 & 0.997 & 45 & 45.913 \\

\midrule
RoBERTa-mimic-DeCLUTR & 0.538 & 1 & 317.300 \\
 & 0.617 & 3 & 54.242 \\
 & 0.699 & 24 & 26.233 \\
 & 0.742 & 50 & 44.607 \\

\midrule
RoBERTa-mimic-note & 0.959 & 3 & 115.873 \\
 & 0.967 & 8 & 127.278 \\
 & 0.973 & 23 & 56.543 \\
 & 0.977 & 34 & 38.244 \\



\bottomrule

\end{tabular}

\caption{Quantitative analysis of graph properties for different models. Cosine similarity thresholds are derived from the 90th, 95th, 98th, and 99th percentiles. The number of components reflects the count of connected components given the threshold chose, and the average degree of the top $N_t$ refers to the graph degree across the top $N_t$ connected components (here $N_t$ is chosen to be 3).}
\label{tab:network-analysis}
\end{table*}

The network analysis highlights that whilst the cosine distances between embeddings for the LLMs that used a contrastive loss during pre-training are greater, they retain a similar graph network structure to other LLMs. The number of components (sub-graphs) can be considered a good indicator of the diversity of the whole graph and correlates strongly with the cosine similarity threshold.

\section{Discussion}

\subsection{General}

The pre-training of LLMs is a crucial step in the production of a useful embedding space for downstream tasks which rely on sentence or document level embeddings, such as sequence classification, document information retrieval, clustering and semantic search. We evaluated various approaches to pre-training LLMs on several downstream sequence classification tasks in frozen and full fine-tuned settings.  

We showed that models pre-trained with contrastive loss functions tended to outperform other pre-training approaches across all our domain-specific datasets, with fewer training samples required to obtain reasonable evaluation performance. Whilst the performance in the frozen setting did not match that of the full fine-tuned setting, the pre-training had a clear influence on how \textit{usable} these LLMs embedding spaces are for the downstream tasks where classification boundaries are important. An attempt to integrate further \textit{structured} metadata based on the \textit{note category} information did not seem to provide a performance gain in the classification tasks above standard MLM training. However in the embedding space analysis, the approach yielded better clustering metrics. The metadata for each dataset differed in nature, and the usefulness of this information for the use-cases presented in this work are difficult to determine and whether or not they hold utility for other tasks would require further investigations.

The DeCLUTR based models appear to produce an embedding space quite distinct to the other pre-training methods, with large separation between different documents and classes. However, it is also clear the the cosine similarity of documents within the same class appears low, potentially highlighting that DeCLUTR did not align well with known classes (recall DeCLUTR is unsupervised and had no access to class labels). The network analysis highlighted a surprising consistency in resultant graph spaces: the pattern of node numbers, component sizes, and degree of major components remained quite stable across each LLM, regardless of the apparent differences in cosine distances.

The utility of the embeddings produced by LLMs as direct features for downstream applications is particularly sought after in resource-constrained settings, where further fine-tuning can be difficult. Domain adaptation of open LLMs to the clinical domain within the UK remains an important goal, with our results showing the largest performance gaps between the general LLMs and the UK NHS dataset trained LLMs presented.  

The resource efficiency of domain adaptation through pre-training is not straight-forward. Of course, the most resource friendly approach would be to include no domain adaption at all, although we suggest the potential performance gain offered by all continued pre-training approaches presented here is worth the relatively low cost: all pre-training could be completed in a matter of hours on a single GPU. 


\subsection{Limitations}

\paragraph{DeCLUTR sampling}
 The sampling parameters that produced the best model for each dataset studied were not equal, were dependent on the length of the documents, and were not flexible to varying lengths of documents. Further, the sampling procedure also eliminates documents that are not long enough to satisfy a pre-determined minimum length. This in fact means the DeCLUTR models were trained on a restricted sub-sample when compared to the other pre-training methods.

\paragraph{Note category pre-training loss}
The number of possible contrastive loss functions applicable to our dataset is very large, and we were not able to explore different setups nor vary the hyperparameters extensively. Similarly, differential selection of samples to use for the pre-training objective was not explored in this work. Further work could also utilise modern transformer architecture adaptations to allow larger batch sizes and context windows \cite{muennighoff_generative_2024, longformer2020, dao_flashattention_2022}.


\paragraph{Strict training settings}
We opted to showcase different pre-training methods in a restricted resource setting, with a single low-end GPU and a limit on training time in line with similar budget-oriented LLM training research \cite{geiping_cramming_2022}. For this reason, we have not explored fully the extremes of these approaches and their application to a wider range of datasets and tasks.

\subsection{Conclusion}

This study underscores the significance of pre-training methodology in domain aligning language models with downstream tasks reliant on document representations. Contrastive, self-supervised objectives prove most effective across the sequence classification tasks, outperforming masked language modeling. While incorporating structured metadata during pre-training did not further improve performance, unsupervised methods like DeCLUTR yield more distinct, clustered embedding spaces. Notably, model embedding graph connectivity patterns persist irrespective of pre-training differences, implying consistent high-level structure. 

Domain adaptation to UK NHS data remains critical, with specialized models substantially improving over general ones. The resource efficiency of this adaptation is non-trivial; no adaptation maximizes efficiency without forfeit to performance gains. The low resource approaches presented still confer valuable improvements worthy of their marginal cost.

While we assessed a range of pre-training schemes and NHS-adapted models on subsequent classification performance, open questions persist. Future work should explore modern architectures and objectives, optimized sampling for contrastive learning, enhanced use of metadata, and applications beyond classification like information retrieval. Broader hyperparameter tuning may unveil further gains. Still, this research validates the utility of pre-trained healthcare language models, provides pre-training best practices, and motivates specialized adaptation - advancing practice while illuminating areas for additional inquiry.

\input{acknowledgements}
\input{funding}

\input{contributions}


\bibliographystyle{unsrtnat}
\bibliography{references.bib, all_references.bib}

\appendix


\section{Dataset details}\label{appendix:dataset-details}


\subsection{Document lengths}
\label{appendix:document-lengths}

Distributions and median document lengths for each dataset are provided below in Fig.\ref{appendix:combined-token-hist}. The MIMIC-III dataset has consistently longer documents, with the PSIR dataset having relatively short documents.

\begin{figure}[htp]
    \centering
    \includegraphics[scale=0.4]{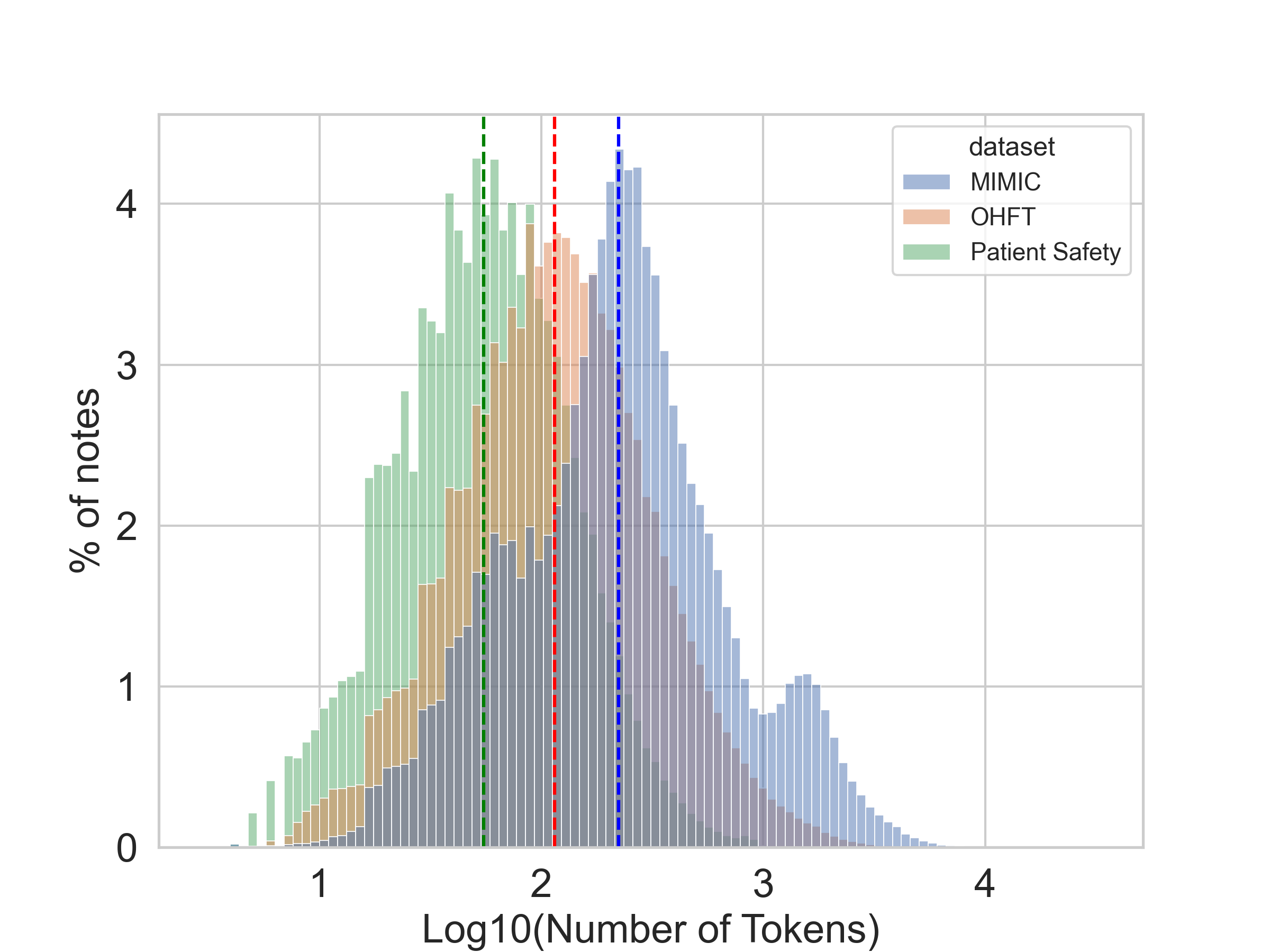}
    \caption{Distribution of the number of tokens in documents for each respective dataset}
    \label{appendix:combined-token-hist}
\end{figure}

\section{Hardware details}
\label{appendix:hardware}
Due to the varying secure locations and computational infrastructure of the hosts of the datasets, it was impossible to match the hardware used. However, all training was carried out on one GPU only. The only difference between the setups was the exact model of GPU. For the PSIR dataset both training and inference were performed on a private single machine hosted by Microsoft Azure with the following main specifications: 1 x NVIDIA Tesla T4 GPU. The OHFT dataset utilised the same Tesla T4 GPU architecture but accessed via a private Amazon Web Services (AWS) instance. The MIMIC-III dataset was housed locally, and we utilised single NVIDIA RTX 2080.

\section{Downstream classification performance} 
\label{appendix:downstream-performance}

\subsection{Patient Safety Incident Reports}

The evaluation results for the severity classification task (P-Sev) and incident type (P-Cat) task are presented in Table \ref{tab:psir-cls-all-results}.

For both tasks we present results in both the frozen LLM and fine-tuned settings. The frozen LLM setting means the parameters of the LLM are set to \textit{not require gradient}, which means gradients will not be computed during the backward pass and keeping these weights fixed and only the newly introduced parameters of the classification head are updated during training. Fine-tuning on the other hand will update all parameters, including the LLM.

\begin{table*}[htp]

\centering
\footnotesize
\begin{subtable}[t]{\textwidth}
\centering
\begin{tabular}{llrrrrr}
\toprule
Model & LLM & Accuracy & F1 & AUC & Precision & Recall \\
\midrule
RoBERTa-base & Frozen & 0.644 & 0.643 & 0.718 & 0.643 & 0.644 \\
RoBERTa-PSIR & Frozen & 0.704 & 0.687 & 0.787 & 0.679 & 0.704 \\
\textbf{RoBERTa-PSIR-DeCLUTR} & Frozen & \textbf{0.786} & \textbf{0.765} & \textbf{0.869} & \textbf{0.752} & \textbf{0.786} \\
RoBERTa-PSIR-note & Frozen & 0.699 & 0.678 & 0.773 & 0.669 & 0.699 \\
\midrule \midrule
RoBERTa-base & Finetuned & 0.847 & 0.636 & 0.926 & 0.640 & 0.646 \\
RoBERTa-PSIR & Finetuned & 0.866 & 0.835 & 0.942 & 0.816 & 0.866 \\
\textbf{RoBERTa-PSIR-DeCLUTR} & Finetuned & \textbf{0.870} & \textbf{0.850} & \textbf{0.944} & \textbf{0.836} & \textbf{0.870} \\
RoBERTa-PSIR-note & Finetuned & 0.863 & 0.833 & 0.936 & 0.814 & 0.863 \\
\bottomrule
\end{tabular}
\caption{Severity classification task (P-Sev)}
\label{subtab:severity-classification}
\end{subtable}
\hfill
\begin{subtable}[t]{\textwidth}
\centering
\begin{tabular}{llrrrrr}
\toprule
Model & LLM & Accuracy & F1 & AUC & Precision & Recall \\
\midrule
RoBERTa-base & Frozen & 0.197 & 0.163 & 0.794 & 0.314 & 0.197 \\
RoBERTa-PSIR & Frozen & 0.498 & 0.474 & 0.851 & 0.494 & 0.499 \\
\textbf{RoBERTa-PSIR-DeCLUTR} & Frozen & \textbf{0.580} & \textbf{0.549} & \textbf{0.900} & \textbf{0.559} & \textbf{0.580} \\
RoBERTa-PSIR-note & Frozen & 0.421 & 0.354 & 0.847 & 0.415 & 0.421 \\
\midrule \midrule
RoBERTa-base & Finetuned & 0.646 & 0.636 & 0.926 & 0.640 & 0.646 \\
RoBERTa-PSIR & Finetuned & 0.665 & 0.652 & 0.935 & 0.655 & 0.665 \\
\textbf{RoBERTa-PSIR-DeCLUTR} & Finetuned & \textbf{0.670} & \textbf{0.659} & \textbf{0.935} & \textbf{0.667} & \textbf{0.670} \\
RoBERTa-PSIR-note & Finetuned & 0.660 & 0.647 & 0.933 & 0.652 & 0.660 \\
\bottomrule
\end{tabular}
\caption{Incident category classification task (P-Cat)}
\label{subtab:incident-category-classification}
\end{subtable}
\caption{Evaluation metrics for the incident category classification task (P-Cat) after one epoch for various models in both frozen and full fine-tuned settings}
\label{tab:psir-cls-all-results}
\end{table*}

\subsection{MIMIC-III}

Evaluation of the different LLMs for the MIMIC-III note category task (M-Cat) and for the MIMIC-III ICD-9 triage task (M-Tri) are presented Table \ref{tab:mimic-all-cls-results}.

\begin{table*}[htp]

\centering
\footnotesize
\begin{subtable}[t]{\textwidth}
\centering
\begin{tabular}{llrrrrr}
\toprule
Model & LLM & Accuracy & F1 & AUC & Precision & Recall \\
\midrule
RoBERTa-base & Frozen & 0.433 & 0.347 & 0.949 & 0.406 & 0.433 \\
RoBERTa-mimic & Frozen & 0.448 & 0.353 & \textbf{0.979} & 0.434 & 0.448 \\
\textbf{RoBERTa-mimic-DeCLUTR} & Frozen & \textbf{0.778} & \textbf{0.770} & 0.962 & \textbf{0.791} & \textbf{0.778} \\
\midrule \midrule
RoBERTa-base & Finetuned & 0.979 & 0.979 & 0.999 & 0.980 & 0.979 \\
RoBERTa-mimic & Finetuned & 0.978 & 0.978 & 0.999 & 0.979 & 0.978 \\
\textbf{RoBERTa-mimic-DeCLUTR} & Finetuned & \textbf{0.985} & \textbf{0.985} & 0.999 & \textbf{0.986} & \textbf{0.985} \\
\bottomrule
\end{tabular}
\caption{MIMIC-III note category task (M-Cat)}
\label{subtab:mimic-note-category}
\end{subtable}
\hfill
\begin{subtable}[t]{\textwidth}
\centering
\begin{tabular}{llrrrrr}
\toprule
Model & LLM & Accuracy & F1 & AUC & Precision & Recall \\
\midrule
RoBERTa-base & Frozen & 0.702 & 0.264 & 0.778 & 0.381 & 0.293 \\
RoBERTa-mimic & Frozen & 0.315 & 0.188 & 0.871 & 0.435 & 0.303 \\
\textbf{RoBERTa-mimic-DeCLUTR} & Frozen & \textbf{0.776} & \textbf{0.545} & \textbf{0.893} & \textbf{0.561} & \textbf{0.542} \\
RoBERTa-mimic-note & Frozen & 0.591 & 0.409 & 0.834 & 0.439 & 0.473 \\
\midrule \midrule
RoBERTa-base & Finetuned & 0.909 & 0.827 & 0.984 & 0.808 & 0.855 \\
RoBERTa-mimic & Finetuned & 0.912 & 0.824 & 0.990 & 0.797 & 0.884 \\
\textbf{RoBERTa-mimic-DeCLUTR} & Finetuned & \textbf{0.917} & \textbf{0.831} & \textbf{0.990} & \textbf{0.794} & \textbf{0.890} \\
RoBERTa-mimic-note & Finetuned & 0.906 & 0.819 & 0.987 & 0.788 & 0.867 \\
\bottomrule
\end{tabular}
\caption{MIMIC-III ICD-9 triage task (M-Tri)}
\label{subtab:mimic-triage}
\end{subtable}
\caption{Evaluation metrics for text classification tasks after one epoch}
\label{tab:mimic-all-cls-results}
\end{table*}

\subsection{OHFT}
Evaluation results for the OHFT note category task (O-Cat) and OHFT Accepted Triage Team task (O-Tri) are presented in Table \ref{tab:ohft-all-cls-results}.

\begin{table*}[htp]

\centering
\footnotesize
\begin{subtable}[t]{\textwidth}
\centering
\begin{tabular}{llrrrrr}
\toprule
Model & LLM & Accuracy & F1 & AUC & Precision & Recall \\
\midrule
RoBERTa-base & Frozen & 0.212 & 0.171 & 0.624 & 0.156 & 0.212 \\
RoBERTa-OHFT & Frozen & 0.224 & 0.163 & 0.697 & 0.162 & 0.224 \\
\textbf{RoBERTa-OHFT-DeCLUTR} & Frozen & \textbf{0.292} & \textbf{0.255} & \textbf{0.709} & \textbf{0.272} & \textbf{0.292} \\
\midrule \midrule
RoBERTa-base & Finetuned & 0.380 & 0.348 & 0.800 & 0.453 & 0.380 \\
\textbf{RoBERTa-OHFT} & Finetuned & \textbf{0.455} & \textbf{0.431} & \textbf{0.852} & \textbf{0.463} & \textbf{0.455} \\
RoBERTa-OHFT-DeCLUTR & Finetuned & 0.404 & 0.390 & 0.821 & 0.434 & 0.404 \\
\bottomrule
\end{tabular}
\caption{OHFT note category task (O-Cat)}
\label{subtab:ohft-note-category}
\end{subtable}
\hfill
\begin{subtable}[t]{\textwidth}
\centering
\begin{tabular}{llrrrrr}
\toprule
Model & LLM & Accuracy & F1 & AUC & Precision & Recall \\
\midrule
RoBERTa-base & Frozen & 0.424 & 0.390 & 0.752 & 0.415 & 0.424 \\
RoBERTa-OHFT & Frozen & 0.495 & 0.469 & 0.830 & 0.572 & 0.495 \\
\textbf{RoBERTa-OHFT-DeCLUTR} & Frozen & \textbf{0.557} & \textbf{0.539} & \textbf{0.831} & \textbf{0.565} & \textbf{0.557} \\
RoBERTa-OHFT-note & Frozen & 0.414 & 0.391 & 0.767 & 0.466 & 0.414 \\
\midrule \midrule
RoBERTa-base & Finetuned & 0.677 & 0.665 & 0.918 & 0.692 & 0.677 \\
\textbf{RoBERTa-OHFT} & Finetuned & \textbf{0.752} & \textbf{0.753} & \textbf{0.935} & \textbf{0.769} & \textbf{0.752} \\
RoBERTa-OHFT-DeCLUTR & Finetuned & 0.738 & 0.738 & 0.934 & 0.750 & 0.738 \\
RoBERTa-OHFT-note & Finetuned & 0.744 & 0.743 & 0.933 & 0.745 & 0.744 \\
\bottomrule
\end{tabular}
\caption{OHFT Accepted triage task (O-Tri)}
\label{subtab:ohft-triage}
\end{subtable}
\caption{Evaluation metrics for text classification tasks after one epoch}
\label{tab:ohft-all-cls-results}
\end{table*}

\section{Performance of open clinical models}

In Table. \ref{tab:other-model-results-first-epoch} we provide some further evaluation results of similarly sized open LLMs pre-trained on biomedical and clinical text, including models trained on all MIMIC-III notes. BioLinkBERT and Bio-ClinicalBERT have previously achieved near state of the art when applied to open medical NLP datasets \cite{huang2019clinicalbert, BioBERT, alsentzer_publicly_2019}. We have results for only the MIMIC-III triage task (M-Tri) and the OHFT accepted triage task (O-Tri). We can see a slight performance drop with the open LLMs when compared with our own on the M-Tri task, and a larger performance drop for the OHFT dataset. This highlights the relative importance of domain pre-training for the UK based dataset.

\begin{table*}[htp]
\centering
\footnotesize
\begin{subtable}{\textwidth}
\centering
\begin{tabular}{llrrrrr}
\toprule
Model & LLM & Accuracy & F1 & AUC & Precision & Recall \\
\midrule
BioLinkBERT-base & Frozen & 0.492 & 0.298 & 0.877 & 0.386 & 0.413 \\
Bio-ClinicalBERT & Frozen & 0.615 & 0.405 & 0.820 & 0.435 & 0.440 \\
\midrule \midrule
BioLinkBERT-base & Finetuned & 0.942 & 0.870 & 0.988 & 0.846 & 0.904 \\
Bio-ClinicalBERT & Finetuned & 0.919 & 0.827 & 0.988 & 0.812 & 0.855 \\
\bottomrule
\end{tabular}
\caption{MIMIC-III ICD-9-triage task (M-Tri)}
\label{subtab:mimic-icd9-triage}
\end{subtable}
\hfill
\begin{subtable}{\textwidth}
\centering
\begin{tabular}{llrrrrr}
\toprule
Model & PLM & Accuracy & F1 & AUC & Precision & Recall \\
\midrule
BioLinkBERT-base & Frozen & 0.376 & 0.356 & 0.706 & 0.410 & 0.376 \\
Bio-ClinicalBERT & Frozen & 0.395 & 0.382 & 0.713 & 0.435 & 0.395 \\
\midrule \midrule
BioLinkBERT-base & Finetuned & 0.703 & 0.700 & 0.915 & 0.714 & 0.703 \\
Bio-ClinicalBERT & Finetuned & 0.715 & 0.714 & 0.922 & 0.718 & 0.715 \\
\bottomrule
\end{tabular}
\caption{OHFT accepted triage team task (O-Tri)}
\label{subtab:ohft-accepted-triage}
\end{subtable}
\caption{Evaluation metrics for text classification tasks after one epoch}
\label{tab:other-model-results-first-epoch}
\end{table*}

\section{Cluster analysis}
A common approach to exploring a dataset through an LLMs embedding space is to perform a form of unsupervised clustering analysis. A simple analysis using the K-Means clustering technique is provided below. The dataset used here is the MIMIC-III ICD-9 triage task (M-Tri). The following clustering metrics are reported: David Bouldin index (DBi) \cite{davies_cluster_1979}, Calinski Harabaz Index (CHi), and the silhouette score (SS). 

The DBi determines the average similarity measure of each derived cluster with its most similar cluster, with similarity defined as the ratio of within-cluster distances to between-cluster distance (far apart clusters with little dispersion resulting in a better, lower score). The CHi is the ratio of the sum of between-cluster and within-cluster dispersion, with higher scores indicating more separable clusters. The SS assesses the overlap of clusters and ranges from -1 to 1 with 1 being optimal. 

We find that the model trained on MIMIC-III using MLM only (RoBERTa-mimic) appears to perform best, with DeCLUTR models actually fairing worse. This is a little surprising considering the DeCLUTR models generally were optimal for downstream classification adaptation. 

\begin{table*}[htp]
\footnotesize
\centering
\begin{tabular}{lrrrr}
\toprule
Model & DBi score (<) & CH score (>) & Silhouette score (>) & Optimal K \\
\midrule
RoBERTa-base & 2.103 & 478.670 & 0.147 & 5 \\
DeCLUTR-base & 3.031 & 284.237 & 0.096 & 4 \\
RoBERTa-base-DeCLUTR & 2.848 & 300.310 & 0.114 & 4 \\
RoBERTa-mimic & 1.663 & 911.911 & 0.235 & 4 \\
RoBERTa-mimic-note & 1.582 & 755.344 & 0.230 & 6 \\
\bottomrule
\end{tabular}
\caption{K-means cluster analysis results for each pre-trained LLM related to the Mimic-III ICD-9-Triage dataset.}
\label{appendix:cluster-analysis}
\end{table*}

\section{Pre-training effects on token classification}

One aspect of this work was to determine how the \textit{pre-training} effected the document level embeddings, however it is also interesting to consider how the word level embeddings have changed: in particular we may expect the contrastive loss functions to have moved the LLMs objective function away from the original MLM objective which should impact token level tasks. 

We investigated the performance of each pre-trained model from the Mimic-III dataset on three token classification tasks (these were readily available with gold standard ground truths). The tasks were Named Entity Recognition (NER) tasks formed during various I2B2 challenges \cite{uzuner_2010_2011, sun_evaluating_2013, stubbs_automated_2015}. 

Micro averaged F1 scores for each task are provided in Table \ref{tab:token-cls-results}, and to determine the utility of the models embeddings as features we explore downstream fine-tuning with the LLM frozen and compare with full fine-tuning. Results do not show any major differences between the models, and generally keeping the LLM frozen did not allow any learning of the task in two of the three tasks. In the fine-tuned setting, all models converge on similar performance in line with other studies utilising the same tasks \cite{rohanian_lightweight_2023}.

\begin{table*}[htp]

\centering
\footnotesize
\begin{tabular}{lcccc}
\toprule
Model  & PLM & i2b2 2010 & i2b2 2012 & i2b2 2014 \\
\midrule
RoBERTa-base & Frozen & 0.052 & 0.042 & 0.639  \\
DeCLUTR-base & &0.043 & 0.065 & 0.651 \\
RoBERTa-base-DeCLUTR & & 0.083 & 0.092 & 0.551 \\
RoBERTa-mimic & & 0.041 & 0.08 & 0.636 \\
RoBERTa-mimic-note & & 0.006 & 0.067 & 0.616  \\

\midrule \midrule
RoBERTa-base & Fine-tuned & 0.847 & 0.83 & 0.975 \\
DeCLUTR-base & & 0.844 & 0.833 & 0.976 \\
RoBERTa-base-DeCLUTR & &0.854 & 0.837 & 0.977  \\
RoBERTa-mimic & & 0.854 & 0.847 & 0.976  \\
RoBERTa-mimic-note &  & 0.855 & 0.841 & 0.979  \\

\bottomrule
\end{tabular}

\caption{Token classification results. For brevity we report F1 micro for each of the models for each dataset}
\label{tab:token-cls-results}
\end{table*}

\section{Training details}\label{appendix:training-details}

\paragraph{Data splits}
In order to avoid direct data leakage from the continued language model pre-training steps and the subsequent downstream classification tasks, we created entirely separate subsets of data: one partition formed the full training and validation sets for the pre-training and the other formed the entire training and validation sets for the downstream tasks. For the different language model pre-training approaches, we subset approximately 250,000 notes for each our datasets. 

\paragraph{Pre-processing}

For language modelling with transformer based models minimal data cleaning is required, as the tokenization of inputs paired with the \textit{contextualised} representations of words means we want to keep as much of the original input as possible. Typical processing steps would include removal of carriage returns, tabs, and white space and any poorly encoded characters.

\paragraph{Hyperparameter choices}
\label{sec:experiments:hyperparas}
Table \ref{tab:hyperparas} reports the main hyperparameter choices for our experiments. Notably, the batch size has direct bearing on the difficulty of the contrastive objective the DeCLUTR and note contrastive models were trained with. In contrastive loss paradigms, for any given anchor, the model needs to make the correct match out of the $N_b$ reports in the batch. The probability of successfully finding the correct match by chance decreases with the batch size. We set the batch size to 32 on the single-GPU machine as this was the maximum possible across all span length options we tried. 

The original authors of DeCLUTR found 2 anchors was optimal for training, whereas the number of positives had little effect. The optimal span length differed between each of our datasets: 1024 for MIMIC-III, 64 for OHFT and Patient Safety, which appears in part related to the average length of document per dataset.

For the note contrastive models, we found combining MLM and the note category losses to be optimal, although when given equal weighting the note category loss often dominated due to it being \textit{new} to the already MLM pre-trained model. Thus, we found lowering the weighting of that loss function dramatically improved the subsequent downstream performance. Whilst we did not explore this thoroughly, we found a weighting of 0.1 for the note category loss performed reasonably.
Each of the different experiments and objectives can require different hyperparameters and we opted to follow those used by original implementations where possible.

\begin{table*}[h]

\centering
    \footnotesize
    \begin{tabular}{l l l l l l}
    \addlinespace
    Parameter & MLM & DeCLUTR & Note Contrastive & Downstream tasks \\ \toprule
    Batch size & 16 & 32 & 16 & 16 \\ 
    Gradient accumulation steps & 4 & 1 & 1 & 1  \\
    Embedding dimension & 768 & 768 & 768  & 768 \\
    Learning rate & $1\times10^{-5}$ & $1\times10^{-5}$ & $1\times10^{-5}$& $1\times10^{-4}$ \\
    Optimiser & Adam W & Adam W & Adam W & Adam W \\
    Span length & - & [16, 64, 512, 1024] & - & - \\ 
    Contrastive loss weight & - & - & [0.1,0.3, 1.0] & - \\
    Epochs & 3 & 3 & 3 & 5 \\    
    \bottomrule
    \end{tabular}
    
    \caption{Overview of hyperparameters used in our experiments. All training regimes utilised a linear scheduler with warm-up.}
    \label{tab:hyperparas}
\end{table*}

\section{DeCLUTR Sampling}\label{appendix:declutr-sampling}

The DeCLUTR sampling algorithm used enforces a minimum length of documents dependent on the span length, and the number of anchor spans to derive. The effect of the minimum length on number of suitable samples for each dataset is provided in Table \ref{tab:DeCLUTR-span-distributions}.
\begin{table}[h]

\centering
    \small
    \begin{tabular}{c   c @{\hspace{1.25\tabcolsep}} @{\hspace{1.25\tabcolsep}} c c @{\hspace{1.25\tabcolsep}}}   
    \addlinespace
    \toprule
    &  \multicolumn{3}{c}{Proportion of 250k LM training dataset} \\
    \cmidrule{1-4}
   Min. document length  & OHFT &  PSIR & MIMIC-III \\ 
   \toprule
        16 & 0.97  & 0.88 & 0.99  \\
        32 & 0.89 & 0.69 & 0.95 \\
        64 & 0.72 & 0.41 & 0.97 \\
        128 & 0.46 & 0.15 & 0.92 \\
        256 &  0.21 & 0.04 & 0.81 \\
        512 &  0.07 & 0.02 & 0.57 \\
        1024 & 0.02 & 0.0 & 0.35\\
        2048 & 0.002 & 0.0 & 0.11\\
    \bottomrule
    \end{tabular}
    
    \caption{Sample distributions for different DeCLUTR sampling minimum document lengths for each of the datasets: OHFT, PSIR, and MIMIC-III. The proportion refers to the amount of samples that meet each of the minimum document length thresholds.}
    \label{tab:DeCLUTR-span-distributions}
\end{table}

\section{DeCLUTR extended results}\label{sec:declutr-extend}
More granular evaluation results for the incident report severity (P-Sev) and incident category (P-Cat) tasks, varying numbers of epochs of pre-training with the DeCLUTR approach (i.e. both the MLM and contrastive loss objective).  The models were trained using documents with a minimum length of 64 tokens, see Table \ref{tab:DeCLUTR-span-distributions} for more details.

\begin{table*}[h]
\centering
\tiny
    \begin{tabular}{c c c @{\hspace{1.\tabcolsep}} c @{\hspace{0.5\tabcolsep}} c @{\hspace{1.\tabcolsep}} c @{\hspace{1.\tabcolsep}} c @{\hspace{1.\tabcolsep}} c @{\hspace{1.\tabcolsep}} c @{\hspace{1.\tabcolsep}} c @{\hspace{1.\tabcolsep}}}
    \addlinespace
    \multicolumn{3}{c}{Model}  & \multicolumn{2}{c}{Performance after 1 epoch} & & & & \\ 
    \cmidrule{1-9} 
    Name & PLM & Pre. Epochs & Trainable params.(m) & Accuracy & ROC AUC & $F_1$  & Prec. & Recall \\ \toprule
        DeCLUTR-base-incident & Frozen &
         2 & 0.5 & 0.758 & 0.842 & 0.725 & 0.713 & 0.758 \\
         & & 3 & 0.5 & 0.763 & 0.847 & 0.719 & 0.709 & 0.763 \\
         & & 5 & 0.5 & 0.763 & 0.848 & 0.75 & 0.741 & 0.763 \\ 
         & & 10 & 0.5 & \textbf{0.771} & 0.855 & 0.752 & 0.74 & \textbf{0.771} \\
         & & 25 & 0.5 & 0.769 & \textbf{0.859} & \textbf{0.756} & \textbf{0.746} & 0.769 \\
         & & 50 & 0.5 & 0.755 & 0.840 & 0.717 & 0.705 & 0.755 \\
        \midrule
        RoBERTa-incident-DeCLUTR & Frozen & 
         2 & 0.5 & 0.775 & 0.858 & 0.750 & 0.736 & 0.775 \\ 
         & & 3 & 0.5 & \textbf{0.786} & 0.869 & 0.765 & 0.752 & \textbf{0.786} \\ 
         & & 5 & 0.5 & 0.780 & \textbf{0.871} & \textbf{0.769} & \textbf{0.760} & 0.780 \\ 
         & & 10 & 0.5 & 0.782 & 0.865 & 0.751 & 0.731 & 0.762 \\
         & & 25 & 0.5 & 0.776 & 0.866 & 0.752 & 0.740 & 0.776 \\
         & & 50 & 0.5 & 0.765 & 0.851 & 0.734 & 0.720 & 0.765 \\ 
        \midrule\midrule 
        DeCLUTR-base-incident & Fine-tuned & 
         2 & 125 & 0.859 & 0.936 & 0.833 & 0.816 & 0.859 \\
         & & 3 & 125 & \textbf{0.861} & \textbf{0.937} & 0.823 & 0.803 & \textbf{0.861} \\
         & & 5 & 125 & 0.858 & 0.936 & 0.841 & 0.828 & 0.858 \\ 
         & & 10 & 125 & 0.859 & 0.934 & 0.827 & 0.808 & 0.859 \\
         & & 25 & 125 & 0.854 & 0.934 & \textbf{0.846} & \textbf{0.835} & 0.854 \\
         & & 50 & 125 & 0.851 & 0.931 & 0.792 & 0.767 & 0.851 \\
        \midrule
        RoBERTa-incident-DeCLUTR & Fine-tuned & 
         2 & 125 & \textbf{0.870} & \textbf{0.944} & \textbf{0.850} & \textbf{0.836} & \textbf{0.870} \\ 
         & & 3 & 125 & 0.868 & 0.942 & 0.838 & 0.819 & 0.868 \\ 
         & & 5 & 125 & 0.867 & 0.941 & 0.831 & 0.811 & 0.867 \\ 
         & & 10 & 125 & 0.858 & 0.938 & 0.825 & 0.807 & 0.858 \\ 
         & & 25 & 125 & 0.863 & 0.939 & 0.833 & 0.815 & 0.863 \\ 
         & & 50 & 125 & 0.857 & 0.934 & 0.831 & 0.815 & 0.857 \\  
         \bottomrule
    \end{tabular}
    \caption{Evaluation metrics for the severity classification task (P-Sev) for various models in both frozen and full fine-tuned settings of DeCLUTR models at various different total epochs of further pre-training for performance after one epoch of training on the task.}
    \label{table:extended-results-severity}
\end{table*}

\begin{table*}[h]
\centering
\tiny
\begin{tabular}{c c c @{\hspace{1.\tabcolsep}} c @{\hspace{0.5\tabcolsep}} c @{\hspace{1.\tabcolsep}} c @{\hspace{1.\tabcolsep}} c @{\hspace{1.\tabcolsep}} c @{\hspace{1.\tabcolsep}} c @{\hspace{1.\tabcolsep}} c @{\hspace{1.\tabcolsep}}}
    \addlinespace
    \multicolumn{3}{c}{Model}  & \multicolumn{2}{c}{Performance after 1 epoch} & & & & \\ 
    \cmidrule{1-9} 
    Name & PLM & Pre. Epochs & Trainable params.(m) & Accuracy & ROC AUC & $F_1$  & Prec. & Recall \\ \toprule
        DeCLUTR-base-incident & Frozen 
        & 2 & 0.5 & 0.548 & 0.878 & 0.511 & 0.525 & 0.548 \\
         & & 3 & 0.5 & 0.542 & 0.883 & 0.503 & 0.509 & 0.542 \\
         & & 5 & 0.5 & 0.559 & 0.886 & 0.527 & 0.531 & 0.559 \\ 
         & & 10 & 0.5 & 0.562 & 0.888 & 0.532 & 0.534 & 0.562 \\
         & & 25 & 0.5 & \textbf{0.577} & \textbf{0.893} & \textbf{0.544} & \textbf{0.557} & \textbf{0.577} \\
         & & 50 & 0.5 & 0.550 & 0.877 & 0.518 & 0.538 & 0.550 \\
        \midrule
        RoBERTa-incident-DeCLUTR & Frozen 
         & 2 & 0.5 & 0.555 & 0.883 & 0.521 & 0.527 & 0.555 \\ 
         & & 3 & 0.5 & 0.568 & 0.892 & 0.531 & 0.541 & 0.568 \\ 
         & & 5 & 0.5 & \textbf{0.580} & \textbf{0.900} & \textbf{0.549} & \textbf{0.559} & \textbf{0.580} \\ 
         & & 10 & 0.5 & 0.578 & 0.894 & 0.544 & 0.551 & 0.578 \\
         & & 25 & 0.5 & 0.574 & 0.892 & 0.545 & 0.545 & 0.574 \\
         & & 50 & 0.5 & 0.572 & 0.883 & 0.548 & 0.552 & 0.572 \\ 
        \midrule\midrule 
        DeCLUTR-base-incident & Fine-tuned 
         & 2 & 125 & 0.648 & 0.930 & 0.641 & 0.651 & 0.648 \\
         & & 3 & 125 & \textbf{0.661} & 0.930 & \textbf{0.651} & \textbf{0.657} & \textbf{0.661} \\
         & & 5 & 125 & 0.649 & \textbf{0.931} & 0.632 & 0.641 & 0.649 \\ 
         & & 10 & 125 & 0.658 & 0.931 & 0.647 & 0.647 & 0.658 \\
         & & 25 & 125 & 0.651 & 0.931 & 0.640 & 0.643 & 0.651 \\
         & & 50 & 125 & 0.645 & 0.929 & 0.631 & 0.635 & 0.645 \\
        \midrule
        RoBERTa-incident-DeCLUTR & Fine-tuned 
        & 2 & 125 & \textbf{0.670} & \textbf{0.935} & 0.659 & 0.667 & \textbf{0.670} \\ 
         & & 3 & 125 & 0.668 & 0.933 & \textbf{0.660} & \textbf{0.672} & 0.668 \\ 
         & & 5 & 125 & 0.660 & 0.934 & 0.646 & 0.660 & 0.660 \\ 
         & & 10 & 125 & 0.658 & 0.932 & 0.644 & 0.653 & 0.658 \\ 
         & & 25 & 125 & 0.659 & \textbf{0.935} & 0.649 & 0.654 & 0.659 \\ 
         & & 50 & 125 & 0.653 & 0.934 & 0.644 & 0.654 & 0.653 \\  
         \bottomrule
    \end{tabular}
    \caption{Evaluation metrics for the incident category classification task (P-Cat) for various models in both frozen and full fine-tuned settings of DeCLUTR models at various different total epochs of further pre-training for performance after one epoch of training on the task.}
\label{table:extended-results-type}
\end{table*}


\begin{table}[htp]
\centering
\footnotesize

\label{tab:fewshot-frozen-results-first-epoch}
\begin{tabular}{llrrrrr}
\toprule
                 & Sample size &   16  &   32  &   64  &   128 &   200 \\
Domain pre-training & Task &       &       &       &       &       \\
\midrule
None & M-Cat & 0.028 & 0.098 & 0.082 & 0.113 & 0.347 \\
     & M-Tri & 0.028 & 0.130 & 0.207 & 0.013 & 0.492 \\
     & O-Cat &   NaN & 0.028 & 0.015 & 0.042 & 0.171 \\
     & O-Tri & 0.071 & 0.109 & 0.095 & 0.186 & 0.079 \\
     & P-Cat & 0.022 & 0.011 & 0.011 & 0.011 & 0.011 \\
     & P-Sev & 0.466 & 0.430 & 0.196 & 0.430 & 0.196 \\

\midrule
MLM & M-Cat & 0.075 & 0.184 & 0.029 & 0.167 & 0.353 \\
     & M-Tri & 0.098 & 0.212 & 0.349 & 0.196 & 0.510 \\
     & O-Cat &   NaN & 0.021 & 0.081 & 0.108 & 0.163 \\
     & O-Tri & 0.069 & 0.134 & 0.086 & 0.168 & 0.213 \\
    & P-Cat & 0.015 & 0.011 & 0.031 & 0.013 & 0.038 \\
     & P-Sev & 0.430 & 0.271 & 0.440 & 0.196 & 0.431 \\

\midrule
DeCLUTR & M-Cat & 0.269 & 0.287 & 0.495 & 0.618 & 0.770 \\
         & M-Tri & 0.146 & 0.095 & 0.315 & 0.431 & 0.683 \\
         & O-Cat & 0.120 & 0.097 & 0.194 & 0.248 & 0.255 \\
         & O-Tri & 0.190 & 0.238 & 0.265 & 0.356 & 0.453 \\
         & P-Cat & 0.024 & 0.031 & 0.020 & 0.065 & 0.119 \\
         & P-Sev & 0.435 & 0.519 & 0.361 & 0.506 & 0.494 \\
         
\midrule
Note contrastive & M-Tri & 0.025 & 0.038 & 0.091 & 0.190 & 0.426 \\
     & O-Tri & 0.106 & 0.212 & 0.173 & 0.349 & 0.413 \\
     & P-Cat & 0.027 & 0.011 & 0.027 & 0.015 & 0.015 \\
     & P-Sev & 0.237 & 0.467 & 0.437 & 0.454 & 0.196 \\
     
\bottomrule
\end{tabular}
\caption{F1 macro score on all tasks after one epoch of training with different number of samples per class. Base LLMs were frozen and only the classification head received updates.}
\end{table}

\begin{table}[htp]
\centering
\footnotesize
\label{tab:fewshot-finetuned-results-first-epoch}
\begin{tabular}{llrrrrr}
\toprule
                 & Sample size &   16  &   32  &   64  &   128 &   200 \\
Domain pre-training & Task &       &       &       &       &       \\
\midrule
None & M-Cat & 0.296 & 0.831 & 0.878 & 0.921 & 0.979 \\
     & M-Tri & 0.050 & 0.023 & 0.748 & 0.732 & 0.827 \\
     & O-Cat & 0.121 & 0.165 & 0.238 & 0.316 & 0.348 \\
     & O-Tri & 0.069 & 0.115 & 0.331 & 0.466 & 0.490 \\
     & P-Cat & 0.011 & 0.012 & 0.011 & 0.057 & 0.397 \\
     & P-Sev & 0.196 & 0.223 & 0.196 & 0.196 & 0.412 \\
     
\midrule
MLM & M-Cat & 0.531 & 0.892 & 0.968 & 0.981 & 0.978 \\
     & M-Tri & 0.282 & 0.367 & 0.596 & 0.802 & 0.824 \\
     & O-Cat & 0.223 & 0.245 & 0.367 & 0.388 & 0.431 \\
     & O-Tri & 0.221 & 0.236 & 0.494 & 0.669 & 0.706 \\
     & P-Cat & 0.022 & 0.012 & 0.017 & 0.443 & 0.557 \\
     & P-Sev & 0.451 & 0.435 & 0.404 & 0.442 & 0.197 \\
     
\midrule
DeCLUTR & M-Cat & 0.734 & 0.921 & 0.967 & 0.981 & 0.985 \\
     & M-Tri & 0.294 & 0.493 & 0.727 &   NaN & 0.831 \\
     & O-Cat & 0.237 & 0.247 & 0.307 & 0.328 & 0.390 \\
     & O-Tri & 0.218 & 0.380 & 0.493 & 0.504 & 0.605 \\
     & P-Cat & 0.051 & 0.037 & 0.099 & 0.399 & 0.452 \\
     & P-Sev & 0.204 & 0.450 & 0.464 & 0.391 & 0.401 \\
     
\midrule
Note contrastive& M-Tri & 0.134 & 0.258 & 0.537 & 0.759 & 0.827 \\
                 & O-Tri & 0.232 & 0.210 & 0.409 & 0.539 & 0.618 \\
                 & P-Cat & 0.013 & 0.012 & 0.019 & 0.129 & 0.519 \\
                 & P-Sev & 0.430 & 0.196 & 0.430 & 0.340 & 0.196 \\
                 
\bottomrule
\end{tabular}
\caption{F1 macro score on all tasks after one epoch of training with different number of samples per class. All models were fully fine-tuned.}
\end{table}

\section{Ablation results}
In order to determine the effect of the contrastive loss components of the \textit{note category} pre-training, we investigated isolating each of the MLM and note category losses to create two separate models. Due to data access constraints, we perform this analysis only on  the MIMIC-III and OHFT datasets. 

For the MIMIC-III ICD-9 triage task (M-Tri) we found that the MLM only models generally perform similarly to the combined loss model, with only a $0.05$ drop in F1 macro in the frozen setting and a $0.1$ increase in the full fine-tuned setting. However, the note category loss only model affects downstream task performance dramatically, with a $0.2$ and $0.1$ drop in F1 macro in the frozen and fine-tuned settings respectively. 

With the OHFT Accepted triage team task (O-Tri), there was very little difference: note only loss lead to a drop of $0.05$ and $0.01$ in frozen and fine-tuned settings respectively. The MLM only loss model had a subtle increase of $0.02$ and $0.01$ in frozen and fine-tuned settings, but this is negligible.  

The drop in performance for the MIMIC-III task may be related to the relationship between the note category meta data and the subsequent downstream tasks: the ICD-9 triage task (M-Tri), as it actually utilises only one type of clinical note, \textit{discharge summaries} and thus the influence of other note types is not examined. 

\end{document}

%% file: acknowledgements.tex
\section{Acknowledgements}
The authors would like to thank the members of the Patient Safety Team in NHS England who engaged with us throughout the project and shared their in-depth knowledge of the area to help shape our exploration. We would also like to acknowledge the work and support of the Oxford Research Informatics Team: Tanya Smith, Research Informatics Manager, Adam Pill, Suzanne Fisher, Research Informatics Systems Analysts and Lulu Kane Research Informatics Administrator. 

%% file: funding.tex
\section{Funding}
NT was supported by the EPSRC Center for Doctoral Training in Health Data Science (EP/S02428X/1) and completed part of this work during a PhD internship with NHS England in late 2022. AK, ANH, and DWJ were supported in part by the NIHR AI Award for Health and Social Care (AI-AWARD02183). DWJ is part supported by an NIHR Infrastructure Programme (NIHR203316).

%% file: contributions.tex
\section*{Contributions}
N.T, D.S, A.K, and A.N.J conceptualised this work. N.T and D.S curated the datasets. N.T developed pre-processing, experiment running and analysis code. N.T. performed experiments across the three datasets. D.S performed extension experiments on PSIR data. N.T and D.S explored and evaluated experimental results. N.T drafted the manuscript. D.S, D.W.J, A.K, and A.N.H revised and edited the manuscript. All authors read and approved the final version of the manuscript.